\documentclass[letterpaper]{article} 
\usepackage[preprint]{aaai2027}  
\usepackage[hyphens]{url}  
\usepackage{graphicx} 
\usepackage{natbib} 
\usepackage{caption} 
\usepackage{booktabs}
\usepackage{amsmath,amssymb}
\usepackage{microtype}
\usepackage{listings}
\title{An Exact Generate--Transform Decomposition of Small-LLM Team Scaling\\
Across Orchestration Architectures}
\author{Blaz Bertalanic, Carolina Fortuna}
\affiliations{Jo\v{z}ef Stefan Institute, Ljubljana, Slovenia}

\begin{document}
\maketitle

\begin{abstract}
Replacing one LLM agent with a collaborating team can raise accuracy, but whether scaling the team helps, and which architecture to scale, is unclear. Sweeping eight agent orchestration architectures across five instruction-tuned 7--9B models, five short-answer benchmarks, and an executable-code benchmark up to 30 calls, we find that the returns to team scaling are sharply task-dependent: from three to thirty calls accuracy rises by up to 17 points on the two arithmetic word-problem benchmarks (GSM8K, GSMHard) but by at most four on ARC, GPQA, and MMLU, for every architecture, a split the usual task-averaged number conceals.  Proposer-Critic captures the arithmetic gains, scaling steepest and, in aggregate, surpassing every other architecture at the largest budget (item-clustered intervals exclude zero), though it ranks among the weakest elsewhere, and no architecture wins across tasks.

We explain these trajectories with an exact generate--transform decomposition.  Partitioning any workflow into proposal coverage and a downstream transform, any accuracy change splits exactly into an \emph{extensive} coverage dividend and an \emph{intensive} transformation change.  The decomposition diagnoses each task: arithmetic offers coverage headroom that a critic-guided transform converts, whereas the multiple-choice benchmarks either saturate in coverage or fail to convert it, and on open-ended code generative recovery nearly vanishes so accuracy tracks coverage.  At equal call budgets token cost still varies 2.1$\times$.  Extra calls therefore create candidate opportunity that only some architectures, on some tasks, convert.  Team scaling is a task- and architecture-specific bet, not a uniform lever.
\end{abstract}

\section{Introduction}

Practitioners increasingly replace one LLM call with a team: agents debate \citep{du2023debate,liang2024divergent}, pass proposals through aggregation layers \citep{wang2024moa}, or communicate over learned graphs \citep{zhuge2024graphs}.  A team is also a compute allocation: the extra calls can generate independent candidates, critique them, revise a single answer serially, or synthesize intermediate reports.  Choosing among these options matters: our largest workflows make 30 calls per problem, and architectures at that same call budget differ 2.1$\times$ in observed tokens.  Yet most evidence fixes one team size and one base model while changing graph, roles, and synthesis together.  A favorable result at five calls on one model says little about whether the design will keep improving at 30, remain cost-effective, or transfer to another model or task.

The gap is increasingly consequential because adaptive systems already select collaboration modes, roles, models, or workflows per query \citep{yue2025masrouter,zhang2025maas,yu2026adaptorch}.  They are optimizing over primitives whose scaling behavior is still poorly characterized.  A further call can act at two stages: it can make a correct candidate available, or it can help later computation preserve, repair, and synthesize the evidence already present.  Conflating the two hides which stage limits scaling.  Where scaling helps depends sharply on the task, and the architecture that gains most on one task can be among the weakest on another (Section~\ref{sec:res}).

We divide each workflow into two functional stages.  The \emph{generate} stage is the first tier that emits candidate answers.  The \emph{transform} stage contains later critics, refiners, synthesizers, and the final manager. We ask three questions in a common experimental protocol.  \emph{First (Q1)}, when and where does scaling a team help, and does the answer depend on the task?  \emph{Second (Q2)}, which architecture, if any, captures the gains as the budget grows?  \emph{Third (Q3)}, do those gains arise because more calls cover more candidate answers, or because later calls transform the available information more effectively? We make three contributions:
\begin{enumerate}
\itemsep0pt
\item Across eight architectures, five small LLMs, and six benchmarks, we show that the returns to team scaling (\textit{Q1}) are sharply task-dependent: scaling from three to thirty calls lifts accuracy by up to 17 points on the arithmetic word-problem benchmarks (GSM8K, GSMHard) but by at most four on ARC, GPQA, and MMLU, a split that task-averaged reporting hides, and Proposer-Critic is the architecture that captures the arithmetic gains while ranking among the weakest elsewhere.
\item We formulate an exact generate--transform decomposition that explains these trajectories.  Unlike fixed-pool selection bounds, in which an uncovered item is lost, it retains generative recovery and so applies to hierarchical managers that synthesize answers outside the proposal pool.  It separates the value of added coverage from covered-case success and recovery, and diagnoses, task by task, whether scaling is limited by missing coverage or by a transform that fails to convert it. \textit{It can inform where and how to cut}.
\item Across the full sweep, the decomposition reveals that no architecture (\textit{Q2}) wins over models and tasks, and equal-budget token cost varies 2.1$\times$.  The decomposition attributes the pervasive early saturation to a transformation term (\textit{Q3}) that is compositional rather than a within-item decline: added coverage falls on harder items that convert weakly, a pattern we confirm on both budget scaling and a controlled prompt-only intervention.
\end{enumerate}
Section~\ref{sec:rw} summarizes related work, Section~\ref{sec:framework} elaborates on the architecture and decomposition, and Section~\ref{sec:method} provides the experimental protocol, while Section~\ref{sec:res} analyzes the results. Finally, Section~\ref{sec:conclusions} concludes the paper.
\section{Related Work}
\label{sec:rw}

\paragraph{Debate, layers, and graphs.}
Multi-agent debate repeatedly exposes agents to peer answers \citep{du2023debate}, and divergent roles can alter the resulting behavior \citep{liang2024divergent}.  Mixture-of-Agents instead passes generations through layers to an aggregator \citep{wang2024moa}, while GPTSwarm treats agent workflows as optimizable computational graphs \citep{zhuge2024graphs}.  Self-consistency and Self-Refine are useful limiting cases: independent proposal aggregation and serial revision, respectively \citep{wang2023selfconsistency,madaan2023selfrefine}.

The closest broad comparisons are MultiAgentBench, which includes star, chain, tree, and graph coordination protocols on interactive scenarios \citep{zhu2025multiagentbench}, and the information-propagation study of \citet{shen2025topologies}, which analyzes error and correct-information diffusion as graph sparsity changes.  \citet{kim2025scalingagents} standardize tools, prompts, and compute across five canonical tool-using agent architectures and derive predictive principles (capability, overhead, redundancy, error propagation) at a fixed scale, rather than a full budget sweep with an exact coverage-versus-transformation accounting.  Adaptive systems select collaboration modes, roles, models, or workflows per query \citep{yue2025masrouter,zhang2025maas,yu2026adaptorch}.  These works motivate architecture selection.  Our complementary goal is complete budget trajectories for fixed small-model workflows, direct accounting between their first answer-producing tier and final output, and an exact decomposition of how the two stages contribute to scaling.

\paragraph{Scaling and diversity.}
MacNet organizes more than a thousand agents in DAGs and fits logistic performance curves, finding that topology affects the curve \citep{qian2025macnet}.  \citet{yang2026agentscaling} instead derive architecture-agnostic information bounds and an effective-channel count, showing that heterogeneous channels can substitute for many homogeneous agents.  In LLM evaluation panels, correlated errors likewise reduce nine nominal judges to roughly two effective votes \citep{kohli2026effectivevotes}.  These results establish that nominal agent count is not informational count.  Repeated-sampling coverage laws \citep{brown2024monkeys} and voting-based call scaling \citep{chen2024morecalls} characterize this generate stage in isolation.  We add a common small-model budget sweep across role-asymmetric workflows and account for proposal generation and subsequent transformation separately.  A single topology-wide effective team size is ambiguous here, because critics, refiners, and synthesizers produce dependent intermediate answers, so we use agreement only as a manipulation check in the matched Star/Persona-Star comparison and rely on transfer events for cross-architecture analysis.

\paragraph{Oracle bounds, selection, and generative aggregation.}
Candidate-pool oracles are established upper bounds for model selection.  SelectLLM, for example, analyzes the gap between a learned selector and an oracle over available models \citep{maurya2025selectllm}.  LLM-Blender makes the adjacent distinction between ranking candidates and generatively fusing them \citep{jiang2023llmblender}.  Concurrent fixed-pool work further separates recoverable mass, selection-signal quality, and harm to correct outputs \citep{hu2026oraclegap}.  If the system must select from a fixed pool, $O=0$ implies $Y=0$.  Hierarchical managers can instead synthesize an answer absent from the initial pool.  Generative Self-Aggregation has already demonstrated such successes when every sampled answer is wrong \citep{li2025gsa}.  Aggregation Fine-Tuning and Recursive Self-Aggregation likewise combine parallel proposal generation with sequential synthesis \citep{li2026aft,venkatraman2025rsa}.  We use the proposal boundary as a common diagnostic and jointly measure coverage, loss, and recovery as heterogeneous fixed workflows scale.

The selection-versus-synthesis distinction is close to the selection bottleneck identified by \citet{maryanskyy2026selection}, who show that generator diversity pays only when the downstream selector is sufficiently reliable.  Homogeneous debate also exhibits consensus collapse, in which aggregation discards correct answers already generated \citep{bertalanic2026consensus}.  Our accounting places such discard and generative repair in one exact identity and evaluates both across parallel and serial workflows.

\paragraph{Compute-normalized evaluation.}
Extra agents also mean extra test-time compute.  Under matched reasoning-token budgets, single-agent systems can match or exceed several multi-agent designs on multi-hop reasoning \citep{tran2026equaltokens}.  We therefore report both requested calls and observed tokens.

\section{Orchestration Architectures and Decomposition  Framework}
\label{sec:framework}

We assume an orchestration  architecture divides its node budget between two functions: creating candidate answers and transforming the resulting evidence.  Let $N$ be the requested node budget, equivalently the team size and the number of LLM calls per problem, and $L$ the number of nodes in the first tier that emits a parseable candidate answer.  Planners instructed not to answer, along with all later critics, refiners, and managers, are excluded from $L$.

\subsection{Orchestration architectures}
\label{sec:architectures}

\begin{figure*}[t]
\centering
\includegraphics[width=0.82\textwidth]{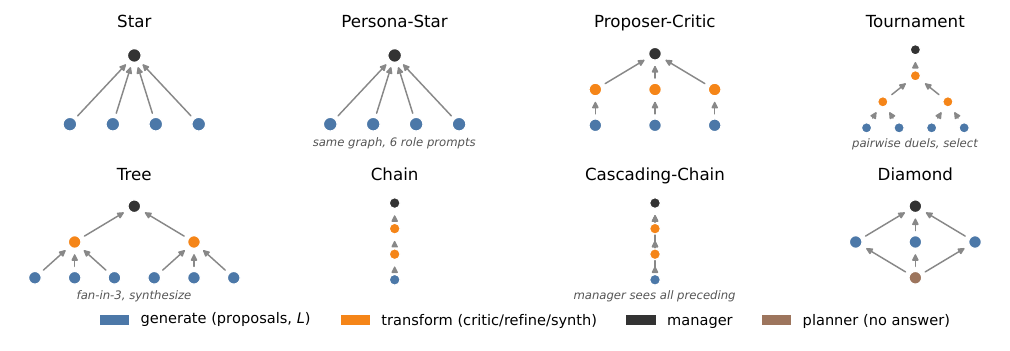}
\caption{The eight orchestration architectures, nodes colored by functional stage: generate (the $L$ proposals), transform (critics, refiners, synthesizers, duelists), the manager, and Diamond's non-answering planner.  \textsc{Tournament} selects through pairwise duels, \textsc{Tree} merges via fan-in-three synthesis, and \textsc{Persona-Star} shares Star's graph but differs in prompts.  Counts are illustrative.}
\label{fig:topologies}
\end{figure*}

We evaluate eight directed acyclic graphs (Figure~\ref{fig:topologies}), each terminating in an active manager and grouped by how the first answer-producing tier of size $L$ scales with the budget.
\begin{itemize}
\itemsep0pt
\item \textbf{Parallel, proposal-expanding ($L=N-1$).} \textsc{Star} routes $N-1$ unconstrained workers to a manager.  \textsc{Persona-Star} keeps Star's graph and manager fixed while cycling six reasoning-method prompts (forward, backward, decomposition, step-back, conservative, and contrarian) across workers \citep{wei2022cot,zhou2023leasttomost,zheng2023stepback}, so only its roles differ, and all six are present from $N=7$.
\item \textbf{Hierarchical and filtered ($1<L<N-1$ for $N\ge5$).} \textsc{Proposer-Critic} pairs proposals with critics ($L=\lceil(N-1)/2\rceil$), \textsc{Tournament} filters proposals through a pairwise bracket, \textsc{Tree} merges them with fan-in-three synthesizers, and \textsc{Diamond} routes $\max(1,\lfloor N/5\rfloor)$ non-answering planners to plan-conditioned solvers ($L=N-\max(1,\lfloor N/5\rfloor)-1$).
\item \textbf{Serial depth ($L=1$).} \textsc{Chain} passes one evolving solution through $N-2$ single-parent refiners.  \textsc{Cascading-Chain} lets each refiner read up to five preceding reports.
\end{itemize}

Requested budgets are $N\in\{2,3,5,7,10,15,20,30\}$.  Diamond starts at $N=3$.  Tournament uses the largest complete pairwise bracket within the request, so its actual counts are 9, 19, and 29 at requested 10, 20, and 30, and all other conditions use the requested count.  Equal call budgets do not buy equal proposal budgets: at $N=30$ the first tier holds $L=29$ proposals for both stars, 15 for \textsc{Proposer-Critic} and \textsc{Tournament}, 22 for \textsc{Tree}, 23 for \textsc{Diamond}, and 1 for both chains.  The remaining calls are critics, duelists, synthesizers, refiners, planners, and the manager. Exact node compositions are tabulated in Appendix~A.

\subsection{Proposal coverage and agreement}

For valid proposal answers $a_1,\ldots,a_{L_v}$, exact-answer pairwise agreement is $A_{\rm pair}=\binom{L_v}{2}^{-1} \sum_{i<j}\mathbf{1}[a_i=a_j]$.  Agreement is undefined when $L_v<2$ and is used only when the proposal boundary is held fixed, principally in the Star/Persona-Star intervention.  Role-asymmetric architectures are instead compared by loss, recovery, and final accuracy below.

\subsection{Generate--Transform decomposition}
\label{sec:gentr}
Let $O=1$ when at least one proposal is correct and $Y=1$ when the final answer is correct.  A final answer can then cross the proposal boundary in either direction: downstream computation can discard an available correct answer or recover when every proposal is wrong.  At a given budget, define the corresponding unconditional masses:
\begin{align}
\ell &= P(O=1,Y=0) &&\text{(transfer loss)},\\
r &= P(O=0,Y=1) &&\text{(transfer recovery)}.\label{eq:transfer}
\end{align}
These two events give the exact endpoint identity
\begin{equation}
P(Y=1)-P(O=1)=r-\ell.
\label{eq:identity}
\end{equation}
Consequently, the oracle gap $P(O=1)-P(Y=1)$ (proposal coverage minus final accuracy) is $\ell-r$, a \emph{net} quantity, not the rate at which the final agent discards a correct proposal.  A selector restricted to the proposal pool has $r=0$.  Recovery can be nonzero here because critics, refiners, and managers are generative reasoners rather than passive voting rules.

To analyze scaling, write $O_N=P(O=1)$ (proposal coverage, the accuracy of an oracle over the proposal pool) and $Y_N=P(Y=1)$ at budget $N$, and define
\begin{equation}
s_N=P(Y=1\mid O=1), g_N=P(Y=1\mid O=0).
\end{equation}
Here $s$ is covered-case success (one minus conditional discard), $g$ is generative recovery, and $v_N=s_N-g_N$ is proposal leverage: the accuracy lift associated with an available correct proposal.  Total probability gives the response identity:
\begin{equation}
Y_N=O_Ns_N+(1-O_N)g_N=g_N+v_NO_N.
\label{eq:level-law}
\end{equation}
For any two conditions $A$ and $B$ (here, two team sizes of a single architecture), write $\Delta x=x_B-x_A$ and $\bar x=(x_A+x_B)/2$.  Applying the exact midpoint product identity to Eq.~\ref{eq:level-law} yields:  
\begin{align} 
\label{eq:two-margin} 
\Delta Y &= \underbrace{\bar v\Delta O}_{\mathcal E: \ \substack{\text{extensive coverage} \\ \text{dividend}}} 
&\quad+ \underbrace{\Delta g+\bar O\Delta v}_{\mathcal I: \ \substack{\text{intensive} \\ \text{transformation change}}}  
\end{align}

Thus $\Delta Y=\mathcal E+\mathcal I$ exactly, without a fitted functional form.  We call Eq.~\ref{eq:two-margin} the two-margin \emph{Generate--Transform decomposition}: it is an exact accounting identity, not an empirical power curve.  Algebraically it is a symmetric rate--composition decomposition \citep{kitagawa1955components}.  Our instantiation uses proposal coverage as the composition and the conditional success rates as the rate, keeping generative recovery inside the accounting rather than assuming a selector.  $\mathcal E$ weights the coverage change by midpoint leverage $\bar v$, so it credits only coverage the downstream stage can exploit.  $\mathcal I=\bar O\Delta s+ (1-\bar O)\Delta g$ measures transformation change at fixed midpoint opportunity weights.  Because the final answer is the manager's own output rather than a tally over the workers, $s$ is the rate at which the manager returns an available correct answer and $1-s$ the rate at which it discards one.  A negative $\mathcal I$ records a fall in aggregate covered-case success.  The paired coverage transitions below attribute this either to a same-item decline or to harder newly-covered items entering the covered population, and we find it is chiefly the latter.

For a paired intervention $A\to B$, marginal conditional rates can still be misleading because $A$ and $B$ need not cover the same items.  We therefore partition paired trials by their realized coverage transition $(O_A,O_B)\in\{00,01,10,11\}$.  With $\pi_{ij}=P(O_A=i,O_B=j)$,
\begin{equation}
\Delta Y=\sum_{i,j\in\{0,1\}}\pi_{ij}
E[Y_B-Y_A\mid O_A=i,O_B=j].
\label{eq:paired-transition}
\end{equation}
Each term is the stratum's exact contribution to the paired effect.  Rate differences have long been separated into composition and conditional-rate components.  Here the paired runs make the coverage transitions directly observable.

For QA we additionally recompute deterministic plurality over the logged proposals using the experiment's item-seeded tie break.  This offline control invokes no LLM and distinguishes voting from active synthesis.

\section{Experimental Protocol}
\label{sec:method}

\paragraph{Models and tasks.}
We evaluate Llama-3.1-8B-Instruct, Ministral-3-8B-Instruct-2512, NVIDIA-Nemotron-Nano-9B-v2, Qwen2.5-7B-Instruct, and Qwen3-8B.  Nemotron and Qwen3 run with thinking disabled, keeping the five models in a common standard-inference setting.  We use QA as shorthand for five short-answer benchmarks (a small answer space, unlike open-ended code): ARC-Challenge (1,165 items), GPQA (198), the arithmetic sets GSM8K (1,319) and GSMHard (1,017), and a 724-item hard-subject MMLU subset \citep{clark2018arc,rein2024gpqa,cobbe2021gsm8k,gao2023pal, hendrycks2021mmlu}.  Functional code uses all 164 HumanEval problems scored with HumanEval+ tests \citep{chen2021humaneval,liu2023evalplus}.

\paragraph{Inference protocol.}
Each cell has three runs with distinct prescribed node seeds.  vLLM serves each model \citep{kwon2023vllm}.  Sampling uses temperature 0.4 and nucleus probability $p=0.95$.  Maximum generation is 1,024 tokens on QA and 2,048 on code. Each parent report is capped at 120 tokens on QA and 512 on code.  The final answer is always the manager's parsed output.  HumanEval+ candidates execute in the EvalPlus sandbox.  Prompts and parsing rules are included in the code and summarized in Appendix~B.

\paragraph{Statistics.}
The analyzed sweep contains 4,179,735 QA and 154,980 code team trials, representing 48,498,195 and 1,798,260 individual agent responses.  We average the three runs within item and use 1,000 item-clustered bootstrap replicates for 95\% intervals.  Paired comparisons retain shared item and run identifiers.  Unless stated otherwise, aggregate numbers weight the 25 QA model--task cells equally.  Invalid (unparseable) proposals are rare, with an equal-cell mean of 0.026\%.  They are excluded from agreement calculations but remain part of the requested-call cost.

Compatible logs provide a direct $N=1$ condition for all 15 hard-QA model--task cells (GPQA, GSMHard, and hard-subject MMLU across five models) with one run, and for all five HumanEval+ models with three runs. It uses the ordinary task prompt without a persona, peer report, or architecture-specific role.  A one-run long-reasoning control instead uses an extended-verification prompt and raises the cap from 1,024 to 10,240 tokens.  It covers 15/15 hard-QA cells. Because prompt and cap change together and generation can stop early, this is a 10$\times$-cap control, not a pure token effect.  Two dense-communication controls run for three rounds and aggregate by plurality: full-mesh Debate, in which every agent sees all peers each round, and a no-peer Self control, in which each agent revises only its own previous answer.  Both cover the 15 hard-QA cells at $N=30$ and the full HumanEval+ budget sweep.  All controls are excluded from primary-sweep counts.

\section{Results}
\label{sec:res}

In this section we leverage the architecture and decomposition framework from Section \ref{sec:framework}, following the protocol from Section \ref{sec:method} to answer \textit{Q1-Q3}. We first show where team scaling pays off and which architecture captures it  (\textit{Q1}), then read the accuracy--cost dependence across architectures (\textit{Q2}), and finally use the generate--transform decomposition to explain the trajectories (\textit{Q3}).  Endpoint, prompt, and debate controls delimit the interpretation, and a controlled prompt-only intervention is found in Appendix~G.1.

\begin{figure*}[t]
\centering
\includegraphics[width=0.65\textwidth]{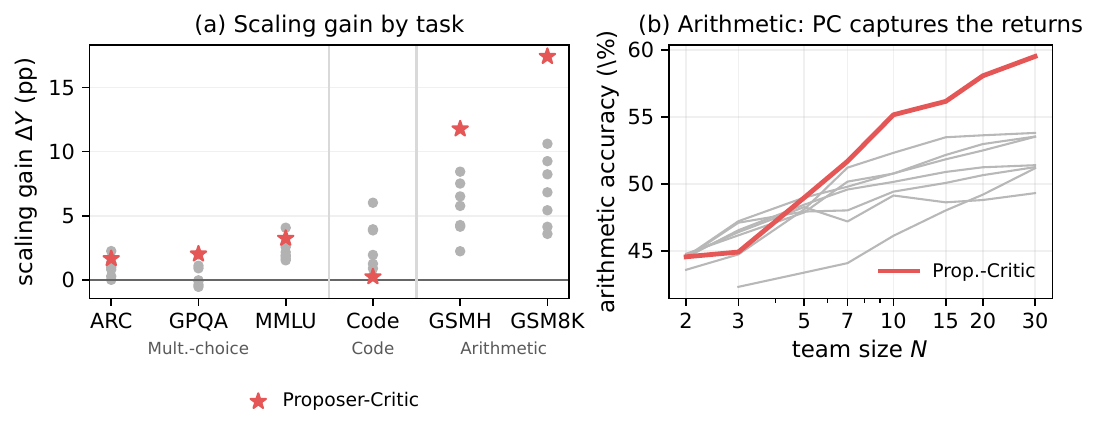}
\caption{Where team scaling pays off.  (a) The scaling gain $\Delta Y$ (the $N=3$ to $N=30$ accuracy change, Section~\ref{sec:gentr}) for every architecture on every benchmark (equal-cell over five models), grouped by task family, with Proposer-Critic marked.  Returns concentrate on arithmetic and are near-zero on ARC, GPQA, and MMLU for all architectures.  (b) Arithmetic accuracy versus team size: Proposer-Critic (red) starts mid-field and separates with scale.}
\label{fig:scaling-by-task}
\end{figure*}

\begin{figure*}[t]
\centering
\includegraphics[width=\textwidth]{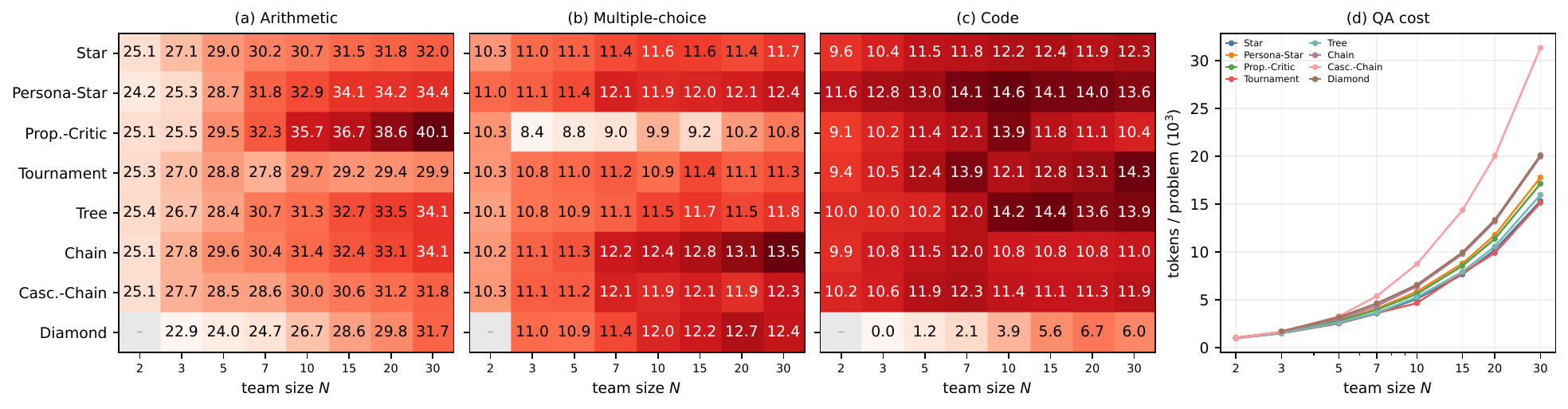}
\caption{Accuracy gain over a single agent (the mean individual worker), by task category: (a) arithmetic, (b) multiple-choice, (c) code, each cell an equal mean over the category's model--task cells with per-panel color scales.  Gains on arithmetic (25 to 40 points) are two to three times those on multiple-choice (8 to 14) or code, and Proposer-Critic tops arithmetic at $N=30$ while trailing on multiple-choice. (d) QA tokens per problem versus team size: cost climbs at architecture-dependent rates to a 2.1$\times$ spread by $N=30$.  Diamond is undefined at $N=2$.}
\label{fig:scaling-cost}
\end{figure*}

\subsection{Scaling returns concentrate on arithmetic}
\label{sec:concentrate}

We find that team scaling does not help uniformly.  Figure~\ref{fig:scaling-by-task}(a)  plots every architecture's accuracy gain from $N=3$ to $N=30$ on each benchmark.  We notice the gains cluster by task: on the two arithmetic word-problem benchmarks (GSM8K, GSMHard) the best architecture adds 12 to 17 points, on open-ended code 4 to 6, and on ARC, GPQA, and MMLU at most about four points for \emph{any} architecture.  A single task-averaged number blends the large arithmetic effect with the near-flat scaling of the others into a misleadingly uniform figure.

Within the tasks that scale, one architecture stands out.  Proposer-Critic is the steepest arithmetic scaler (Figure~\ref{fig:scaling-by-task}(b)): sixth of eight at $N=3$, it overtakes the field by $N\approx10$ and in aggregate at $N=30$ surpasses every other architecture on arithmetic by 5.7 to 10.2 points, all item-clustered 95\% intervals excluding zero (the runner-up margin is $+5.7$, CI $[5.1,6.4]$).  The advantage is modal, not universal: Proposer-Critic leads arithmetic for three to four of the five models, a chain for the rest, and it ranks last or near-last on ARC, MMLU, and code.  The decomposition (introduced in Section~\ref{sec:gentr}) explains the split (Section~\ref{sec:decomp}): on arithmetic Proposer-Critic's transform converts the added coverage ($\mathcal I=+8.1$ on GSM8K, $+3.0$ on GSMHard), whereas on GPQA and MMLU it adds coverage but sheds most of it ($\mathcal I=-5.0$ and $-3.6$), and ARC has little to add ($\mathcal E=+1.4$).
\noindent\fbox{%
\parbox{\dimexpr\linewidth-2\fboxsep-2\fboxrule}{%
\textit{Answer to Q1}: The returns to team scaling are sharply task-dependent. From $N=3$ to $N=30$ accuracy rises by up to 17 points on the arithmetic word-problem benchmarks (GSM8K, GSMHard) and by 4 to 6 on open-ended code, but by at most 4 on the multiple-choice benchmarks (ARC, GPQA, MMLU) for every architecture.%
}%
}

\begin{figure*}[t]
\centering
\includegraphics[width=0.75\textwidth]{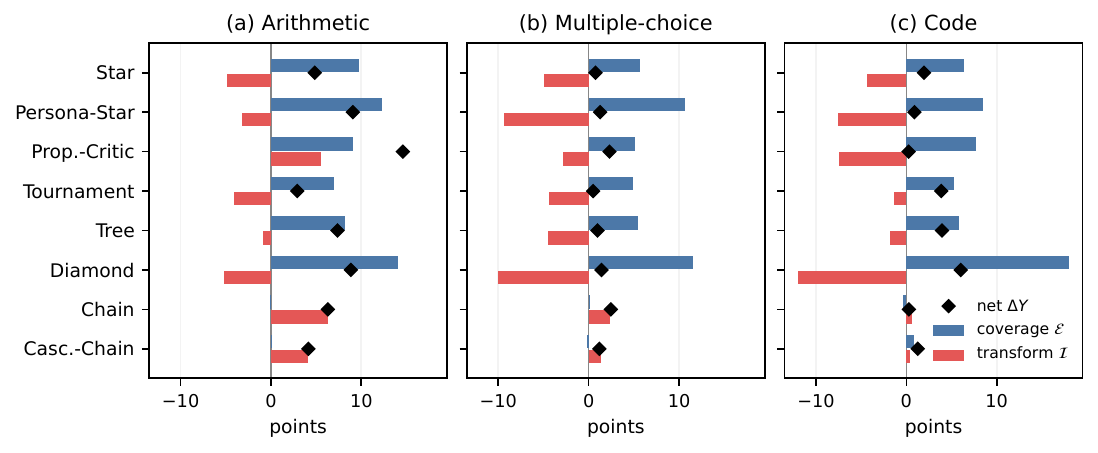}
\caption{Two-margin decomposition of the $N=3\to30$ accuracy change by task category: (a) arithmetic, (b) multiple-choice, (c) code.  Each architecture's change splits into an extensive coverage dividend $\mathcal E$ (blue) and an intensive transformation change $\mathcal I$ (red, extending left when the transform sheds coverage), summing to the net $\Delta Y$ (diamond), equal-cell over each category's model--task cells.  Coverage is added in every category, but the transform converts it only on arithmetic, where Proposer-Critic alone posts a positive $\mathcal I$ and the largest net gain.  On multiple-choice every proposal-expanding design sheds most of its coverage and nets almost nothing, and on code the same shedding, with recovery near zero, leaves the net close to the coverage floor.}
\label{fig:main}
\end{figure*}

\subsection{Scaling is architecture-, model-, and cost-dependent}

Figure~\ref{fig:scaling-cost}(a--c) gives each architecture's gain over a single agent by budget and task category, and the trajectory shape itself is task-dependent.  On multiple-choice (b) and code (c) most of the lift is present by $N=3$ and the curves flatten by $N=10$, whereas on arithmetic (a) they keep climbing to $N=30$, where Proposer-Critic reaches $+40$ points over a single agent (level, not slope).  Over the $N=3\to30$ range Proposer-Critic improves in 21/25 QA model--task cells and Chain in 24/25, the most consistent signs.  Persona-Star and Diamond reach similar aggregate gains through sharply different margin profiles (Section~\ref{sec:decomp}).  We read the $N=3$ to 30 slope, not the intercept, as the scaling result, because the intercept is baseline-sensitive and shrinks against a stronger single agent (Section~\ref{sec:robust}).

Equal team sizes are not equal cost (Figure~\ref{fig:scaling-cost}(d)).  Token cost climbs with $N$ at architecture-dependent rates: at $N=30$ mean QA cost ranges from 15.1k tokens for Tournament to 31.3k for Cascading-Chain, whose accumulating context grows fastest, a 2.1$\times$ spread.  Combining this cost with accuracy (Appendix~E), Tournament, Star, Tree, and Proposer-Critic form the accuracy--cost frontier, with Proposer-Critic highest at 59.1\% and 17.1k tokens.  A purpose-built single-agent control given a 10$\times$ token cap gains 9.18 points over the ordinary single agent and is matched on average by the post-hoc best architecture at 13.5$\times$ the observed tokens, so extra test-time compute in a single agent is itself competitive (Appendix~H).

At $N=30$ the level comparison splits sharply by category (Appendix~E).  On arithmetic the within-model architecture spread is large (mean 14.0 points, up to 25.6 for Llama) and Proposer-Critic leads four of the five models, a chain leading Ministral.  On multiple-choice the spread collapses to a mean of 4.3 points, every design lands within a few points, and a chain is marginally best.  No architecture wins across models in either category.  Model selection stays the larger lever on multiple-choice, where architecture barely moves accuracy, whereas on arithmetic the architecture spread rivals it.  Teams beat a single agent by far more on arithmetic, where one can solve as little as 8\% of items, than on multiple-choice, a like-for-like margin a stronger single agent substantially narrows (Appendix~H).  The HumanEval+ levels tell the same story (Appendix~E): no architecture wins across models, and every design beats a single agent by +6.0 (Diamond) to +14.3 (Tournament) points.

\noindent\fbox{%
\parbox{\dimexpr\linewidth-2\fboxsep-2\fboxrule}{%
\textit{Answer to Q2}: While no architecture wins across all tasks, we find that  \textsc{Proposer-Critic} captures the arithmetic gains, surpassing every other architecture in aggregate at $N=30$ on arithmetic (GSM8K and GSMHard), yet it ranks last or near-last on ARC, MMLU, and code. Equal-budget token cost still varies 2.1$\times$, and \textsc{Proposer-Critic} sits on the accuracy--cost frontier.%
}%
}

\subsection{Scaling gains split into coverage and transformation}
\label{sec:decomp}

Accuracy curves establish whether another call helped, but not what that call purchased.  We apply Eq.~\ref{eq:two-margin} within each model--task cell from $N=3\to30$, average the signed terms, and read them by task category (Figure~\ref{fig:main}).  Added coverage is reliably valuable: across the six proposal-expanding designs (Section~\ref{sec:architectures}) the extensive dividend $\mathcal E$ is positive in every one of the 150 cells, from $+4.9$ to $+14.1$ points depending on category and design.  What separates the tasks is not coverage but whether the downstream transform converts it, and that is the intensive term.

The intensive term $\mathcal I$ has a task-dependent sign (Figure~\ref{fig:main}).  On the multiple-choice benchmarks it is negative for nearly every width design (87 of 90 cells, $-2.8$ to $-10.1$ points) and cancels almost the whole dividend: Diamond turns a $+11.5$ coverage gain into $+1.4$ final points, and no width design nets above $+2.3$.  On arithmetic the transform sheds far less ($\mathcal I$ negative in 43 of 60 cells, and only $-0.9$ to $-5.2$ where it is), and Proposer-Critic reverses it outright ($\mathcal I=+5.55$), converting its coverage and adding recovery for a $+14.6$-point net gain, the largest of any design and category.  Pooled over both QA categories, Proposer-Critic is the only width design whose intensive term does not significantly erode its dividend ($\mathcal I=+0.54$, 95\% CI $[-0.02,1.05]$, Appendix~D).

This intensive decline is consistent with composition rather than a same-item transform decline.  Pairing every item across $N=3$ and $N=30$ by its coverage transition (Eq.~\ref{eq:paired-transition}), on the stratum covered at both budgets final accuracy holds or rises for five of the six width designs ($+2.2$ to $+6.7$ points, with Diamond flat and Tournament $-0.9$), so the drop in aggregate covered-case success is carried by weakly-converting newly-covered items rather than by degradation on shared ones (Appendix~D).

The serial chains invert this in every category.  Their first tier is a single proposal, so $\mathcal E$ is effectively zero and their gains are entirely intensive, driven by recovery: $\mathcal I=+6.35$ (Chain) and $+4.15$ (Cascading-Chain) on arithmetic, $+2.42$ and $+1.33$ on multiple-choice.  An added call therefore has architecture-dependent meaning.  Width buys candidate opportunity and must overcome an intensive headwind, whereas serial depth buys repeated repair of one evolving solution, and nominal team size alone obscures the distinction.

The endpoint masses $\ell$ and $r$ show what the final transformation does with its proposals (full table in the Appendix~D).  Parallel architectures are loss-dominated at $N=30$: Diamond and Persona-Star discard 27.5\% and 25.5\% of covered cases and recover under 5\% of uncovered ones, whereas the chains reverse the pattern, recovering 29.2\% and 24.9\% of uncovered cases against 3\% loss.  Proposer-Critic nearly balances the two.

This contrast shows up directly at the endpoints: the chains recover far more than they lose and finish above their proposal oracle, Proposer-Critic nearly balances the two, and the five other width designs are loss-dominated and finish below their proposal oracle. The crossing boundary that formalizes it, with per-architecture coverage values and its stability across budgets, is given in the Appendix~D. Loss- and recovery-dominance are not fixed architecture labels: the same model and architecture can switch by task.  With Diamond, Qwen3 recovers on 20.8\% of GSM8K trials but only 0.3\% on GPQA, so a one-sided oracle gap would flag only the latter, yet Eq.~\ref{eq:identity} shows both are outcomes of the same transfer stage.

\paragraph{Open-ended code.} The same decomposition (Section~\ref{sec:gentr}) applies, since the proposal boundary needs only executable correctness (a correct program passes the tests) and textual agreement is not used.  The width pattern reproduces on HumanEval+ (Figure~\ref{fig:main}(c)): the six proposal-expanding designs post an extensive dividend of +5.2 to +18.0 points from $N=3$ to 30 offset by a negative intensive term, and Diamond adds the most coverage (proposal oracle +26.1) yet gains only +6.0 final points.  Two differences sharpen the account.  Generative recovery nearly disappears, with $g=P(Y{=}1\mid O{=}0)$ at 0--1\% for Star, Persona-Star, and Diamond against 15--31\% across QA.  And every architecture finishes below its proposal oracle ($Y-O$ from $-1.1$ to $-19.3$), whereas the QA chains finished 12--14 points above.  Open-ended synthesis rarely yields a passing program the pool lacked, so the transform stage preserves or discards coverage but seldom recovers it, and accuracy tracks coverage minus loss.  On this one open-ended benchmark the recovery that lets QA managers exceed their proposal oracle disappears, consistent with a small-answer-space affordance (Appendix~H).
\noindent\fbox{%
\parbox{\dimexpr\linewidth-2\fboxsep-2\fboxrule}{%
\textit{Answer to Q3}: The gains could come from extra calls more often making a correct answer available among the candidates (coverage), or from the final step converting those candidates into a correct final answer. Both matter, but conversion is the bottleneck: coverage rises on every task, yet accuracy improves only where the downstream transform converts it. Conversion succeeds on arithmetic, but on multiple-choice and code the transform discards most of the newly available answers.%
}%
}

\subsection{Choosing an architecture: resolution beats prediction}
\label{sec:selection}

\textbf{(1) Scale only where a single call is weak.} Team scaling adds a lot where one call misses many items and little where it already scores well (Section~\ref{sec:concentrate}): large gains on arithmetic and multi-step reasoning, negligible ones on knowledge and multiple-choice, where a single agent given a longer reasoning budget is competitive at far lower cost (Section~\ref{sec:robust}). If one call already handles your task, do not scale.

\noindent\textbf{(2) If scaling helps, find the architecture by testing, not guessing.} If you can label a small development set, run several architectures at a small budget ($N=10$), keep the best, and deploy it at your target budget ($N=30$), landing within 1.2 points of the best of the eight. If you cannot label anything, deploy the task's historically strongest architecture: Proposer-Critic on arithmetic, a chain otherwise. Rules that route on the decomposition's own signals do no better than this default (1.55, 4.02, and 2.46 points below the best of the eight, versus the default's 1.59), so the decomposition diagnoses scaling but cannot route it (Appendix~D.1).

\subsection{Controls and robustness}
\label{sec:robust}

Against a purpose-built direct $N=1$ call, a stronger single agent using a cleaner prompt (and long reasoning for Nemotron), the margin narrows: every fixed $N=30$ architecture still improves on the 15 hard-QA cells (+3.0 to +6.6), the post-hoc best by 8.9 (13/15, Appendix~H)).

Three-round full-mesh debate beats direct $N=1$ by 8.3 points on the 15 hard-QA cells, but the post-hoc best sparse architecture is higher in 10/15 at 7.3--24.3$\times$ fewer tokens, and the no-peer control edges it by 0.69 point.  On code, where debate runs the full budget sweep, one round of peer exchange captures its entire benefit and beats no-peer revision by 2.4 points, yet debate still only ties the best sparse design at twice the calls.  Dense peer exchange is therefore a costly and task-dependent baseline, not a free win (Appendix~I).

Managers are not deterministic votes: final accuracy exceeds offline plurality by 7.8--13.3 points across the six proposal-expanding architectures, and changing only the manager instruction can cut accuracy through lower recovery (Appendix~G.2).

\section{Limitations}

Except for Star/Persona-Star and Chain/Cascading-Chain, graph and role prompts change together, so we rank orchestration bundles rather than graph structure alone.  Budgets match calls rather than tokens.  We study fixed, homogeneous 7--9B non-thinking teams to $N=30$ on short-answer and executable-code tasks, not frontier or heterogeneous models, tool use, or long-form generation.  The code results use a single benchmark, so the vanished-recovery effect cannot be separated from domain, answer format, or execution-based scoring.

Agreement is answer-space dependent and is used only for the matched Persona intervention.  The Generate--Transform decomposition is exact accounting, not causal mediation or a parametric forecast: $s$ and $g$ condition on populations that can change, so $\mathcal I$ can mix composition and transformation.

\section{Conclusion}
\label{sec:conclusions}
Team scaling is not a uniform lever: its returns concentrate on arithmetic word problems, where Proposer-Critic converts the added coverage at scale, while ARC, GPQA, and MMLU gain little for any architecture.  The generate--transform decomposition makes the difference legible, separating the coverage a workflow adds from whether its transform converts it.  No architecture wins across models and tasks.  The right question is where, and at what budget, scaling pays at all.

\bibliography{refs}

@article{hu2026oraclegap,
  title   = {Oracle Gap and Signal Fidelity: A Fixed-Pool Diagnostic for Test-Time Collaboration},
  author  = {Hu, Jie},
  journal = {arXiv preprint arXiv:2607.17531},
  year    = {2026}
}

@inproceedings{maurya2025selectllm,
  title     = {{SelectLLM}: Query-Aware Efficient Selection Algorithm for Large Language Models},
  author    = {Maurya, Kaushal Kumar and Srivatsa, Kv Aditya and Kochmar, Ekaterina},
  booktitle = {Findings of the Association for Computational Linguistics: ACL 2025},
  pages     = {20847--20863},
  publisher = {Association for Computational Linguistics},
  year      = {2025}
}

@inproceedings{jiang2023llmblender,
  title     = {{LLM-Blender}: Ensembling Large Language Models with Pairwise Ranking and Generative Fusion},
  author    = {Jiang, Dongfu and Ren, Xiang and Lin, Bill Yuchen},
  booktitle = {Proceedings of ACL},
  year      = {2023}
}

@article{li2025gsa,
  title   = {{LLM}s Can Generate a Better Answer by Aggregating Their Own Responses},
  author  = {Li, Zichong and Feng, Xinyu and Cai, Yuheng and Zhang, Zixuan and Liu, Tianyi and Liang, Chen and Chen, Weizhu and Wang, Haoyu and Zhao, Tuo},
  journal = {arXiv preprint arXiv:2503.04104},
  year    = {2025}
}

@inproceedings{li2026aft,
  title     = {The Best of Both Worlds: Combining Parallel and Sequential Inference Scaling via Aggregation Fine-Tuning},
  author    = {Li, Yafu and Wang, Zhilin and Fu, Tingchen and Cui, Ganqu and Yang, Sen and Cheng, Yu},
  booktitle = {Findings of the Association for Computational Linguistics: ACL 2026},
  pages     = {31369--31389},
  publisher = {Association for Computational Linguistics},
  year      = {2026}
}

@article{venkatraman2025rsa,
  title   = {Recursive Self-Aggregation Unlocks Deep Thinking in Large Language Models},
  author  = {Venkatraman, Siddarth and Jain, Vineet and Mittal, Sarthak and Shah, Vedant and Obando-Ceron, Johan and Bengio, Yoshua and Bartoldson, Brian R. and Kailkhura, Bhavya and Lajoie, Guillaume and Berseth, Glen and Malkin, Nikolay and Jain, Moksh},
  journal = {arXiv preprint arXiv:2509.26626},
  year    = {2025}
}

@inproceedings{du2023debate,
  title     = {Improving Factuality and Reasoning in Language Models through Multiagent Debate},
  author    = {Du, Yilun and Li, Shuang and Torralba, Antonio and Tenenbaum, Joshua B. and Mordatch, Igor},
  booktitle = {Proceedings of ICML},
  year      = {2024}
}

@inproceedings{liang2024divergent,
  title     = {Encouraging Divergent Thinking in Large Language Models through Multi-Agent Debate},
  author    = {Liang, Tian and He, Zhiwei and Jiao, Wenxiang and Wang, Xing and Wang, Yan and Wang, Rui and Yang, Yujiu and Tu, Zhaopeng and Shi, Shuming},
  booktitle = {Proceedings of EMNLP},
  year      = {2024}
}

@inproceedings{wang2024moa,
  title     = {Mixture-of-Agents Enhances Large Language Model Capabilities},
  author    = {Wang, Junlin and Wang, Jue and Athiwaratkun, Ben and Zhang, Ce and Zou, James},
  booktitle = {Proceedings of COLM},
  year      = {2024}
}

@inproceedings{zhuge2024graphs,
  title     = {Language Agents as Optimizable Graphs},
  author    = {Zhuge, Mingchen and Wang, Wenyi and Kirsch, Louis and Faccio, Francesco and Khizbullin, Dmitrii and Schmidhuber, J{\"u}rgen},
  booktitle = {Proceedings of ICML},
  year      = {2024}
}

@inproceedings{shen2025topologies,
  title     = {Understanding the Information Propagation Effects of Communication Topologies in {LLM}-based Multi-Agent Systems},
  author    = {Shen, Xu and Liu, Yixin and Dai, Yiwei and Wang, Yili and Miao, Rui and Tan, Yue and Pan, Shirui and Wang, Xin},
  booktitle = {Proceedings of EMNLP},
  year      = {2025}
}

@article{kim2025scalingagents,
  title   = {Towards a Science of Scaling Agent Systems},
  author  = {Kim, Yubin and Gu, Ken and Park, Chanwoo and Park, Chunjong and Schmidgall, Samuel and Heydari, A. Ali and Yan, Yao and Zhang, Zhihan and Zhuang, Yuchen and Liu, Yun and Malhotra, Mark and Liang, Paul Pu and Park, Hae Won and Yang, Yuzhe and Xu, Xuhai and Du, Yilun and Patel, Shwetak and Althoff, Tim and McDuff, Daniel and Liu, Xin},
  journal = {arXiv preprint arXiv:2512.08296},
  year    = {2025}
}

@inproceedings{zhu2025multiagentbench,
  title     = {{MultiAgentBench}: Evaluating the Collaboration and Competition of {LLM} Agents},
  author    = {Zhu, Kunlun and Du, Hongyi and Hong, Zhaochen and Yang, Xiaocheng and Guo, Shuyi and Wang, Zhe and Wang, Zhenhailong and Qian, Cheng and Tang, Xiangru and Ji, Heng and You, Jiaxuan},
  booktitle = {Proceedings of ACL},
  year      = {2025}
}

@inproceedings{yue2025masrouter,
  title     = {{MasRouter}: Learning to Route {LLM}s for Multi-Agent Systems},
  author    = {Yue, Yanwei and Zhang, Guibin and Liu, Boyang and Wan, Guancheng and Wang, Kun and Cheng, Dawei and Qi, Yiyan},
  booktitle = {Proceedings of ACL},
  year      = {2025}
}

@inproceedings{zhang2025maas,
  title     = {Multi-agent Architecture Search via Agentic Supernet},
  author    = {Zhang, Guibin and Niu, Luyang and Fang, Junfeng and Wang, Kun and Bai, Lei and Wang, Xiang},
  booktitle = {Proceedings of ICML},
  year      = {2025}
}

@article{yu2026adaptorch,
  title   = {{AdaptOrch}: Task-Adaptive Multi-Agent Orchestration in the Era of {LLM} Performance Convergence},
  author  = {Yu, Geunbin},
  journal = {arXiv preprint arXiv:2602.16873},
  year    = {2026}
}

@inproceedings{wang2023selfconsistency,
  title     = {Self-Consistency Improves Chain-of-Thought Reasoning in Language Models},
  author    = {Wang, Xuezhi and Wei, Jason and Schuurmans, Dale and Le, Quoc and Chi, Ed and Narang, Sharan and Chowdhery, Aakanksha and Zhou, Denny},
  booktitle = {Proceedings of ICLR},
  year      = {2023}
}

@inproceedings{madaan2023selfrefine,
  title     = {Self-Refine: Iterative Refinement with Self-Feedback},
  author    = {Madaan, Aman and Tandon, Niket and Gupta, Prakhar and Hallinan, Skyler and Gao, Luyu and Wiegreffe, Sarah and Alon, Uri and Dziri, Nouha and Prabhumoye, Shrimai and Yang, Yiming and Welleck, Sean and Majumder, Bodhisattwa Prasad and Gupta, Shashank and Yazdanbakhsh, Amir and Clark, Peter},
  booktitle = {Proceedings of NeurIPS},
  year      = {2023}
}

@inproceedings{wei2022cot,
  title     = {Chain-of-Thought Prompting Elicits Reasoning in Large Language Models},
  author    = {Wei, Jason and Wang, Xuezhi and Schuurmans, Dale and Bosma, Maarten and Ichter, Brian and Xia, Fei and Chi, Ed and Le, Quoc V. and Zhou, Denny},
  booktitle = {Proceedings of NeurIPS},
  year      = {2022}
}

@inproceedings{zhou2023leasttomost,
  title     = {Least-to-Most Prompting Enables Complex Reasoning in Large Language Models},
  author    = {Zhou, Denny and Sch{\"a}rli, Nathanael and Hou, Le and Wei, Jason and Scales, Nathan and Wang, Xuezhi and Schuurmans, Dale and Cui, Claire and Bousquet, Olivier and Le, Quoc and Chi, Ed},
  booktitle = {Proceedings of ICLR},
  year      = {2023}
}

@inproceedings{zheng2023stepback,
  title     = {Take a Step Back: Evoking Reasoning via Abstraction in Large Language Models},
  author    = {Zheng, Huaixiu Steven and Mishra, Swaroop and Chen, Xinyun and Cheng, Heng-Tze and Chi, Ed H. and Le, Quoc V. and Zhou, Denny},
  booktitle = {Proceedings of ICLR},
  year      = {2024}
}

@article{clark2018arc,
  title   = {Think You Have Solved Question Answering? Try {ARC}, the {AI2} Reasoning Challenge},
  author  = {Clark, Peter and Cowhey, Isaac and Etzioni, Oren and Khot, Tushar and Sabharwal, Ashish and Schoenick, Carissa and Tafjord, Oyvind},
  journal = {arXiv preprint arXiv:1803.05457},
  year    = {2018}
}

@inproceedings{rein2024gpqa,
  title     = {{GPQA}: A Graduate-Level Google-Proof Q\&A Benchmark},
  author    = {Rein, David and Hou, Betty Li and Stickland, Asa Cooper and Petty, Jackson and Pang, Richard Yuanzhe and Dirani, Julien and Michael, Julian and Bowman, Samuel R.},
  booktitle = {Proceedings of COLM},
  year      = {2024}
}

@article{cobbe2021gsm8k,
  title   = {Training Verifiers to Solve Math Word Problems},
  author  = {Cobbe, Karl and Kosaraju, Vineet and Bavarian, Mohammad and Chen, Mark and Jun, Heewoo and Kaiser, Lukasz and Plappert, Matthias and Tworek, Jerry and Hilton, Jacob and Nakano, Reiichiro and Hesse, Christopher and Schulman, John},
  journal = {arXiv preprint arXiv:2110.14168},
  year    = {2021}
}

@inproceedings{gao2023pal,
  title     = {{PAL}: Program-Aided Language Models},
  author    = {Gao, Luyu and Madaan, Aman and Zhou, Shuyan and Alon, Uri and Liu, Pengfei and Yang, Yiming and Callan, Jamie and Neubig, Graham},
  booktitle = {Proceedings of ICML},
  year      = {2023}
}

@inproceedings{hendrycks2021mmlu,
  title     = {Measuring Massive Multitask Language Understanding},
  author    = {Hendrycks, Dan and Burns, Collin and Basart, Steven and Zou, Andy and Mazeika, Mantas and Song, Dawn and Steinhardt, Jacob},
  booktitle = {Proceedings of ICLR},
  year      = {2021}
}

@article{chen2021humaneval,
  title   = {Evaluating Large Language Models Trained on Code},
  author  = {Chen, Mark and others},
  journal = {arXiv preprint arXiv:2107.03374},
  year    = {2021}
}

@inproceedings{liu2023evalplus,
  title     = {Is Your Code Generated by {ChatGPT} Really Correct? Rigorous Evaluation of Large Language Models for Code Generation},
  author    = {Liu, Jiawei and Xia, Chunqiu Steven and Wang, Yuyao and Zhang, Lingming},
  booktitle = {Proceedings of NeurIPS},
  year      = {2023}
}

@inproceedings{qian2025macnet,
  title     = {Scaling Large Language Model-based Multi-Agent Collaboration},
  author    = {Qian, Chen and Xie, Zihao and Wang, YiFei and Liu, Wei and Zhu, Kunlun and Xia, Hanchen and Dang, Yufan and Du, Zhuoyun and Chen, Weize and Yang, Cheng and Liu, Zhiyuan and Sun, Maosong},
  booktitle = {Proceedings of ICLR},
  year      = {2025}
}

@article{yang2026agentscaling,
  title   = {Understanding Agent Scaling in {LLM}-Based Multi-Agent Systems via Diversity},
  author  = {Yang, Yingxuan and Qu, Chengrui and Wen, Muning and Shi, Laixi and Wen, Ying and Zhang, Weinan and Wierman, Adam and Gu, Shangding},
  journal = {arXiv preprint arXiv:2602.03794},
  year    = {2026}
}

@article{kohli2026effectivevotes,
  title   = {Nine Judges, Two Effective Votes: Correlated Errors Undermine {LLM} Evaluation Panels},
  author  = {Kohli, Guneet},
  journal = {arXiv preprint arXiv:2605.29800},
  year    = {2026}
}

@article{maryanskyy2026selection,
  title   = {When Agents Disagree: The Selection Bottleneck in Multi-Agent {LLM} Pipelines},
  author  = {Maryanskyy, Artem and Budnikov, Dmitry and Kaliyev, Alibek T.},
  journal = {Applied Sciences},
  volume  = {16},
  number  = {10},
  pages   = {4914},
  doi     = {10.3390/app16104914},
  year    = {2026}
}

@article{kitagawa1955components,
  title   = {Components of a Difference Between Two Rates},
  author  = {Kitagawa, Evelyn M.},
  journal = {Journal of the American Statistical Association},
  volume  = {50},
  number  = {272},
  pages   = {1168--1194},
  doi     = {10.1080/01621459.1955.10501299},
  year    = {1955}
}

@article{tran2026equaltokens,
  title   = {Single-Agent {LLM}s Outperform Multi-Agent Systems on Multi-Hop Reasoning Under Equal Thinking Token Budgets},
  author  = {Tran, Dat and Kiela, Douwe},
  journal = {arXiv preprint arXiv:2604.02460},
  year    = {2026}
}

@inproceedings{bertalanic2026consensus,
  title     = {The Cost of Consensus: Isolated Self-Correction Prevails Over Unguided Homogeneous Multi-Agent Debate},
  author    = {Bertalani{\v{c}}, Bla{\v{z}} and Fortuna, Carolina},
  booktitle = {Proceedings of the ACM Conference on AI and Agentic Systems},
  series    = {CAIS '26},
  pages     = {311--329},
  publisher = {Association for Computing Machinery},
  doi       = {10.1145/3786335.3813137},
  year      = {2026}
}

@inproceedings{kwon2023vllm,
  title     = {Efficient Memory Management for Large Language Model Serving with {PagedAttention}},
  author    = {Kwon, Woosuk and Li, Zhuohan and Zhuang, Siyuan and Sheng, Ying and Zheng, Lianmin and Yu, Cody Hao and Gonzalez, Joseph E. and Zhang, Hao and Stoica, Ion},
  booktitle = {Proceedings of SOSP},
  year      = {2023}
}

@article{brown2024monkeys,
  title   = {Large Language Monkeys: Scaling Inference Compute with Repeated Sampling},
  author  = {Brown, Bradley and Juravsky, Jordan and Ehrlich, Ryan and Clark, Ronald and Le, Quoc V. and R{\'e}, Christopher and Mirhoseini, Azalia},
  journal = {arXiv preprint arXiv:2407.21787},
  year    = {2024}
}

@inproceedings{chen2024morecalls,
  title     = {Are More {LLM} Calls All You Need? Towards the Scaling Properties of Compound {AI} Systems},
  author    = {Chen, Lingjiao and Davis, Jared Quincy and Hanin, Boris and Bailis, Peter and Stoica, Ion and Zaharia, Matei and Zou, James},
  booktitle = {Proceedings of NeurIPS},
  year      = {2024}
}

\clearpage
\appendix
\setcounter{table}{0}
\setcounter{figure}{0}
\renewcommand{\thetable}{S\arabic{table}}
\renewcommand{\thefigure}{S\arabic{figure}}
\setcounter{topnumber}{6}
\setcounter{dbltopnumber}{6}
\setcounter{bottomnumber}{4}
\setcounter{totalnumber}{10}
\renewcommand{\topfraction}{0.95}
\renewcommand{\bottomfraction}{0.90}
\renewcommand{\textfraction}{0.05}
\renewcommand{\floatpagefraction}{0.80}
\renewcommand{\dbltopfraction}{0.95}
\renewcommand{\dblfloatpagefraction}{0.80}

\section*{Supplementary Material}
This supplement documents the experimental and statistical protocol and provides the evidence underlying the main paper's three claims: the returns to team scaling are sharply task-dependent (large on the arithmetic word-problem benchmarks, small on the multiple-choice ones, with Proposer-Critic capturing the arithmetic gains), an exact generate--transform decomposition separates proposal coverage from downstream conversion and explains the split, and no architecture wins across models and tasks.  It also reports the direct and long-reasoning single-agent controls, HumanEval+ results, full-mesh debate and no-peer revision comparison, plurality control, and manager-prompt stress test used to delimit those claims.  The audited cell file and all result tables are regenerated from raw logs by versioned analysis scripts.

\section{Architecture Specification}

All eight conditions are directed acyclic workflows and all terminate in one LLM manager.  The manager is an active reasoner instructed to inspect its parent reports and emit a final answer, not cast a deterministic vote.  We call the conditions \emph{orchestration architectures} because graph and role prompt jointly define most of them.

\paragraph{Star and Persona-Star.}
Star has $N-1$ parent-free workers.  Persona-Star changes only those worker prompts and cycles six methods by worker index: forward solving, backward checking, decomposition, step-back abstraction, conservative calibration, and contrarian search.  The node count, edges, sampling settings, manager prompt, and model remain fixed.  At least one instance of every persona is present from $N=7$ onward.

\paragraph{Proposer-Critic and Tournament.}
Proposer-Critic allocates $(N-1)/2$ pairs where possible.  Each critic sees one proposal and is instructed to find errors before answering.  The manager sees critic outputs plus an unpaired proposal when $N-1$ is odd.  Tournament begins with the largest worker count whose full pairwise bracket fits inside $N$.  Each duelist sees two previous reports and selects or repairs one answer.  The final duelist is relabeled as manager.  Requested budgets 10, 20, and 30 therefore use 9, 19, and 29 actual calls.  Other requested budgets in the sweep match exactly.

\paragraph{Tree and serial chains.}
Tree uses branching factor three.  Complete triples feed synthesizers and any orphan worker feeds the manager directly.  Chain begins with one worker.  Each of its $N-2$ refiners sees only the immediately preceding report.  The manager sees the complete sequence.  Cascading-Chain changes one knob: each refiner sees up to the five most recent reports, while its manager still sees the complete sequence.  Both chains therefore have $L=1$ regardless of $N$.  Their additional calls transform one evolving solution rather than add parallel proposals.

\paragraph{Diamond.}
Diamond allocates $\lfloor N/5\rfloor$ planners (at least one).  Planners are explicitly forbidden to emit a final answer and are excluded from $L$.  Remaining non-manager calls are balanced across plans.  These plan-conditioned solvers form the first answer-producing tier.  The manager sees every solver.

\begin{table*}[t]
\centering\small
\begin{tabular}{lrrllr}
\toprule
Architecture & Actual calls & $L$ & Other non-final roles & Manager input & Tiers \\
\midrule
Star & 30 & 29 & none & 29 worker reports & 2 \\
Persona-Star & 30 & 29 & none & 29 persona-worker reports & 2 \\
Proposer-Critic & 30 & 15 & 14 critics & 14 critics + 1 free proposal & 3 \\
Tournament & 29 & 15 & 13 non-final duelists & 2 finalists & 5 \\
Tree & 30 & 22 & 7 fan-in-three synthesizers & 7 syntheses + 1 orphan & 3 \\
Chain & 30 & 1 & 28 one-parent refiners & all 29 preceding reports & 30 \\
Cascading-Chain & 30 & 1 & 28 window-$\leq5$ refiners & all 29 preceding reports & 30 \\
Diamond & 30 & 23 & 6 pre-answer planners & 23 plan-conditioned answers & 3 \\
\bottomrule
\end{tabular}
\caption{Exact architecture composition at requested budget $N=30$.  $L$ is the first answer-producing tier used for proposal metrics.  Tournament rounds down to the largest pairwise bracket within the requested budget.}
\label{tab:composition}
\end{table*}

\section{Prompts, Inference, and Scoring}

Every QA prompt requires a final parse in the form \texttt{FINAL: <answer>} and a confidence.  Workers solve independently.  Critics receive one parent and must identify flaws before committing.  Refiners receive their parent reports and are instructed to verify and correct them.  Synthesizers and managers receive labeled reports, compare reasoning, and solve the question before emitting one final answer.  Duelists compare two reports, planners propose distinct approaches without answering, and plan-solvers receive one plan and execute it.  Parent reports are token-truncated, not character-truncated.  The verbatim templates for every role are reproduced in Appendix~\ref{sec:prompts}.

Sampling temperature is 0.4 and top-$p$ is 0.95.  QA generations are capped at 1,024 tokens with 120 tokens retained per parent report.  HumanEval+ generations are capped at 2,048 tokens with 512 tokens per parent report and a 24,576-token context cap.  A request seed mixes base seed 42, run id, requested budget, topology, item id, tier, and node id.  This prevents identical prompts within a tier from sharing an RNG stream.  Nemotron uses \texttt{/no\_think}, and Qwen3 passes \texttt{enable\_thinking=False}.

The auxiliary long-reasoning control remains a single agent but uses the foundation runner's extended prompt, which requests multiple approaches and verification, and raises the QA generation cap from 1,024 to 10,240 tokens.  Because the instruction and cap both change and decoding can terminate early, we do not interpret it as a pure token-budget treatment.

The manager-prompt stress test keeps Diamond's $N=30$ DAG, upstream prompts, model, sampling settings, and seed recipe fixed and changes only the final manager instruction.  One variant explicitly tallies candidate answers before deciding whether to follow or override the mode.  The other critiques each distinct answer before synthesizing.  Baseline and variants were launched as separate jobs.  Seeded decoding therefore yields closely matched but not byte-identical proposal pools, so we analyze the paired realized runs rather than describe this as frozen-evidence replay.

The primary sweeps were launched as one-model Slurm jobs on NVIDIA H100 80GB GPUs with eight CPU cores and 80GB of host memory per job.  The code package records the Python dependency bounds and the model-specific vLLM launch flags.  Because the original environment was not frozen to exact package patch versions, this part of the computational record is necessarily partial.

MCQ answers are canonicalized to option letters and numeric answers to integer strings before scoring.  HumanEval candidates are parsed as Python functions and executed against both base and HumanEval+ tests.  A candidate must pass both suites.  Textual plurality and answer entropy are not used as code metrics because distinct correct functions are not fungible strings.

\section{Statistical Protocol}

The analyzed primary sweep contains 4,334,715 team trials and 50,296,455 agent responses.  Proposal coverage and agreement statistics use exactly the architecture's declared first answer-producing tier $L$.  Later critics, refiners, and managers are excluded.

For a model--architecture--task--budget cell, each item's three runs are averaged first.  Confidence intervals resample unique items with replacement for 1,000 replicates.  Persona-Star comparisons are paired on shared $(\text{run},\text{item})$ keys before runs are averaged within item.  Aggregate tables weight model--task cells equally rather than letting GSM8K dominate GPQA by item count.  No multiplicity correction is applied to individual intervals.  The principal Persona agreement result is the uniform sign and the fact that all 25 intervals exclude zero.

For the aggregate Persona-Star effects, we use 5,000 fixed-grid stratified replicates.  Each replicate resamples item identities within benchmark and carries all five models and all runs for a sampled question together, thereby preserving cross-model dependence on the same item.  The resulting 25 cell effects are weighted equally.  Common draws are used for coverage, loss, recovery, and final accuracy, so the transfer identity closes in every replicate.  These intervals quantify item uncertainty conditional on the tested model--task grid.  They do not treat five models or five tasks as random samples from wider populations.

For the paired coverage-transition analysis, every shared $(\text{run},\text{item})$ realization is assigned to one of $(O_{\rm Star},O_{\rm Persona})\in\{00,01,10,11\}$.  Its unconditional contribution is the stratum share times the paired final-accuracy difference within that stratum.  The four contributions sum exactly to the overall Persona-Star effect.  Aggregate intervals use the same benchmark-stratified item bootstrap, carrying all runs and five models for a sampled item together.  The strata are observed stochastic realizations, not latent causal types.

The conflict probe is restricted to $N=30$ Star and Persona-Star trials with $O=1$.  Within each model--task--architecture cell it compares proposal agreement between final-answer loss ($Y=0$) and preservation ($Y=1$).  Bootstrap draws resample items and keep all runs of a sampled item together.  This controls proposal availability within an architecture but not item difficulty, and the two architectures cover different item populations.  The probe is therefore explicitly associational and is not used to infer a cross-system manager effect.

For the Diamond manager-prompt stress test, baseline and each variant are paired by model, task, run, and item.  We reconstruct proposal coverage $O$ from the first-tier correctness vector and apply the same loss--recovery identity.  Aggregate intervals use 5,000 benchmark-stratified item-bootstrap replicates, carry a sampled item jointly across all five models, and weight the 25 cells equally.  We also report exact proposal-vector and coverage-status agreement across the separately decoded pairs to delimit the intervention.

For the dense communication controls, round accuracies first average all available runs within item.  Debate and no-peer Self are paired on their common round-2/round-3 item support.  1,000 bootstrap replicates resample items.  Cumulative cost sums each item's mean prompt-plus-output tokens over rounds 1--3 before taking the cell mean.

The direct-baseline audit is separate from the primary sweep.  It retains 9,695 hard-QA rows across all 15 model--task cells (one run per item) and 2,460 HumanEval+ rows (three runs for each of five models).  The condition uses a single agent, one model call, and the ordinary task prompt without persona, peer, or architecture-specific instructions.  Consequently it answers whether collaboration improves over one clean direct generation.  The separate 10$\times$-cap condition provides a stronger single-agent reference on 14 hard-QA cells, but does not exactly match observed team tokens or isolate the effect of tokens from its extended-reasoning instruction.

The endpoint identity is checked by construction on every audited row.  In the notation of the main paper,
\begin{align*}
\text{net transfer} &= r-\ell \\
 &= P(Y=1)-P(O=1).
\end{align*}

\paragraph{Code and data availability.} All prompt templates, workflow implementations, analysis code, and per-cell result files (cell statistics, decomposition terms with closure errors, transfer masses, and bootstrap intervals) are released with the paper, and each table and figure is produced by a named script in the release.

\section{Exact Two-Margin Scaling Decomposition}

Table~\ref{tab:two-margin} expands Figure~4(a,b) of the main paper.  We compute the midpoint decomposition separately in every model--task cell and average the signed terms only afterward.  Consequently, its closure is not an identity that holds only for an aggregate ``representative'' system: every one of the 200 cell rows satisfies $\Delta Y=\mathcal E+\mathcal I$ to numerical precision.  Item-clustered 95\% intervals (1{,}000 replicates) sharpen the sign counts: every width design's extensive term excludes zero above and the five eroding designs' intensive terms exclude zero below, while Proposer-Critic's intensive interval $[-0.02,1.05]$ straddles zero.  A per-cell count (111 of 150 significantly negative) agrees but is uncorrected for multiplicity, so the equal-cell intervals are the primary evidence.

\begin{table*}[t]
\centering\small\setlength{\tabcolsep}{3.5pt}
\begin{tabular}{lrrrrrrrr}
\toprule
Architecture & $\mathcal{E}$ & $\bar O\Delta s$ & $(1-\bar O)\Delta g$ & $\mathcal{I}$ & $\Delta Y$ & $\Delta Y>0$ & $\mathcal{E}>0$ & $\mathcal{I}<0$ \\
\midrule
Star & +7.28 & -2.65 & -2.23 & -4.88 & +2.40 & 17/25 & 25/25 & 22/25 \\
Persona-Star & +11.32 & -5.14 & -1.80 & -6.94 & +4.39 & 20/25 & 25/25 & 23/25 \\
Prop.-Critic & +6.68 & -0.67 & +1.21 & +0.54 & +7.22 & 21/25 & 25/25 & 18/25 \\
Tournament & +5.72 & -3.12 & -1.14 & -4.26 & +1.46 & 18/25 & 25/25 & 22/25 \\
Tree & +6.55 & -2.46 & -0.56 & -3.02 & +3.53 & 20/25 & 25/25 & 21/25 \\
Chain & +0.00 & -0.18 & +4.17 & +3.99 & +3.99 & 24/25 & 14/25 & 2/25 \\
Casc.-Chain & -0.09 & +0.18 & +2.28 & +2.46 & +2.37 & 18/25 & 10/25 & 6/25 \\
Diamond & +12.52 & -5.49 & -2.64 & -8.13 & +4.39 & 22/25 & 25/25 & 24/25 \\
\bottomrule
\end{tabular}
\caption{Exact two-margin decomposition of QA scaling from requested $N=3$ to $N=30$, in equal-cell percentage points.  The intensive term is split into changes in covered-case success and recovery.  Every row satisfies $\Delta Y=\mathcal{E}+\mathcal{I}$ before rounding.  The last three columns count signs across 25 model--task cells.}
\label{tab:two-margin}
\end{table*}

Table~\ref{tab:both-covered} tests whether the negative intensive term is a same-item transform decline or a composition effect of the growing covered population.  We pair every trial across $N=3$ and $N=30$ on (task, run, item), stratify by the coverage transition $(O_3,O_{30})$, and read the stratum covered at both budgets, which holds item difficulty fixed and varies only the tier width the manager faces.  There final accuracy rises for Star, Persona-Star, Proposer-Critic, and Tree, is flat for Diamond, and falls only for Tournament, whose pairwise bracket can discard a correct finalist.  The aggregate fall in covered-case success is therefore dominated by the newly-covered stratum (contribution $c_{01}$): harder items that convert weakly, mirroring the matched Star--Persona-Star intervention (Appendix~\ref{sec:persona-intervention}) rather than a manager that degrades on answers it already had.  The same pattern holds for the $N=10\to30$ contrast.

\begin{table*}[t]
\centering\small\setlength{\tabcolsep}{5.0pt}
\begin{tabular}{lrrrrrrrr}
\toprule
Architecture & $\pi_{11}$ & $Y^{11}_{3}$ & $Y^{11}_{30}$ & $\Delta Y_{11}$ & $\pi_{01}$ & $c_{01}$ & $c_{11}$ & $\Delta Y$ \\
\midrule
Star & 48 & 85.4 & 88.6 & +3.19 & 13 & +1.48 & +1.00 & +2.40 \\
Persona-Star & 49 & 83.7 & 88.8 & +5.10 & 20 & +2.20 & +1.55 & +4.39 \\
Prop.-Critic & 44 & 79.5 & 86.2 & +6.73 & 15 & +2.50 & +1.42 & +7.22 \\
Tournament & 48 & 86.0 & 85.0 & -0.94 & 11 & +1.31 & -0.22 & +1.46 \\
Tree & 48 & 85.8 & 88.0 & +2.23 & 12 & +1.55 & +0.88 & +3.53 \\
Diamond & 45 & 83.6 & 83.6 & -0.01 & 26 & +4.45 & -0.26 & +4.39 \\
Chain & 40 & 91.1 & 91.3 & +0.22 & 4 & +1.27 & -0.14 & +3.99 \\
Casc.-Chain & 40 & 91.4 & 92.0 & +0.55 & 4 & +1.41 & +0.09 & +2.37 \\
\bottomrule
\end{tabular}
\caption{Paired coverage-transition decomposition of QA scaling from $N=3$ to $N=30$, equal-cell means over 25 model--task cells.  Each trial is paired across the two budgets by (task, run, item) and stratified by its coverage transition $(O_3,O_{30})$; match rate is 100\%.  $\pi_{11}$ is the share of trials covered at both budgets and $\pi_{01}$ the newly-covered share (percent).  $Y^{11}_{3}$ and $Y^{11}_{30}$ are final accuracies on the both-covered stratum and $\Delta Y_{11}$ their difference; $c_{01}$ and $c_{11}$ are the strata contributions to the total $\Delta Y$ (points).  The both-covered stratum holds item difficulty fixed, so a non-negative $\Delta Y_{11}$ argues against a same-item transform decline: accuracy there rises for the four proposal-expanding designs Star, Persona-Star, Proposer-Critic, and Tree, is flat for Diamond, and falls only for Tournament.  The negative intensive term of Table~\ref{tab:two-margin} is therefore consistent with the weak conversion of harder newly-covered items rather than the transform discarding answers it previously returned.}
\label{tab:both-covered}
\end{table*}

The intensive split also clarifies how the architectures differ.  For five width designs, both covered-case success and recovery contribute negatively on average.  Proposer-Critic combines a small negative covered-case component with a positive recovery component.  The resulting mean intensive term is positive despite being negative in 18/25 individual cells.  Chain's gain is almost entirely increased recovery, while Cascading-Chain obtains smaller positive contributions from both intensive components.

Table~\ref{tab:transfer} gives the complete endpoint transfer quantities underlying Section~5.3 of the main paper, including opportunity-normalized discard and recovery.  Its columns are equal means of the 25 cellwise rates, so ratios of the displayed aggregate columns need not reproduce the displayed conditional rates.

\begin{table*}[t]
\centering\small\setlength{\tabcolsep}{5.0pt}
\begin{tabular}{lrrrrrrr}
\toprule
Architecture & $O$ & Loss & $P(Y{=}0\mid O{=}1)$ & Recovery & $P(Y{=}1\mid O{=}0)$ & Net & $Y$ \\
\midrule
Star & 61.5 & 10.8 & 19.6 & 5.7 & 12.6 & -5.1 & 56.4 \\
Persona-Star & 69.7 & 16.7 & 25.5 & 4.8 & 11.7 & -11.9 & 57.8 \\
Prop.-Critic & 59.1 & 11.4 & 20.2 & 11.3 & 27.0 & 0.0 & 59.1 \\
Tournament & 59.2 & 10.9 & 21.8 & 7.1 & 15.2 & -3.9 & 55.4 \\
Tree & 60.6 & 10.5 & 19.9 & 7.2 & 15.1 & -3.3 & 57.3 \\
Chain & 44.6 & 3.3 & 10.5 & 17.0 & 29.2 & +13.8 & 58.4 \\
Casc.-Chain & 44.5 & 2.8 & 9.7 & 15.1 & 24.9 & +12.2 & 56.7 \\
Diamond & 70.1 & 18.2 & 27.5 & 4.9 & 13.6 & -13.3 & 56.7 \\
\bottomrule
\end{tabular}
\caption{Proposal-to-team transfer at $N=30$ on QA (equal mean over 25 model--task cells). $O$ and $Y$ are proposal coverage and final accuracy. Loss and recovery are unconditional masses needed by the accounting identity; the conditional columns measure downstream behavior given the corresponding opportunity. Net is recovery minus loss.}
\label{tab:transfer}
\end{table*}

Writing $v=s-g$ exposes the response form $Y=g+vO$.  Subtracting $O$ gives $Y-O=g-(1-v)O$, hence the exact crossing point $O^\star=g/(1-v)=g/(1-s+g)$.  Table~\ref{tab:response-boundary} instead pools the equal-cell joint masses before calculating conditional rates.  This choice makes the displayed response law close exactly and therefore differs slightly from Table~\ref{tab:transfer}.

\begin{table*}[t]
\centering\small\setlength{\tabcolsep}{5.0pt}
\begin{tabular}{lrrrrrrr}
\toprule
Architecture & $O$ & $s$ & $g$ & $v$ & $O^\star$ & $Y$ & $Y-O$ \\
\midrule
Star & 61.5 & 82.4 & 14.9 & 0.675 & 45.8 & 56.4 & -5.1 \\
Persona-Star & 69.7 & 76.0 & 15.9 & 0.601 & 39.9 & 57.8 & -11.9 \\
Prop.-Critic & 59.1 & 80.8 & 27.7 & 0.531 & 59.0 & 59.1 & -0.0 \\
Tournament & 59.2 & 81.6 & 17.3 & 0.642 & 48.4 & 55.4 & -3.9 \\
Tree & 60.6 & 82.7 & 18.2 & 0.645 & 51.3 & 57.3 & -3.3 \\
Chain & 44.6 & 92.7 & 30.7 & 0.619 & 80.7 & 58.4 & +13.8 \\
Casc.-Chain & 44.5 & 93.7 & 27.1 & 0.665 & 81.1 & 56.7 & +12.2 \\
Diamond & 70.1 & 74.0 & 16.2 & 0.578 & 38.5 & 56.7 & -13.3 \\
\bottomrule
\end{tabular}
\caption{Response-law boundary at $N=30$.  Rates pool the equal-cell joint masses, so $Y=g+(s-g)O$ holds exactly for every row.  A generative transformer finishes above its proposal oracle exactly when $O<O^\star=g/(1-s+g)$.  All values are percentages except proposal leverage $v=s-g$.}
\label{tab:response-boundary}
\end{table*}

The boundary is diagnostic rather than a learned forecast, but which side of it a cell falls on can be estimated before the largest budget.  We label each cell by whether $O<O^\star$ at an earlier budget and test that label at $N=30$ without using the intervening budgets.  Table~\ref{tab:boundary-stability} shows agreement rising from 76.0\% at $B=3$ to 93.0\% at $B=7$ and 96.0\% at $B=10$.  This is held-out-budget persistence on the same model--task cells and benchmark items, not a deployable router.  The next subsection tests selection under model- and item-hold-out directly.

\subsection{Selecting an architecture under strict hold-out}

We test whether the two behaviors support a deployable selector under a model- and item-held-out protocol: leave one model out, split each task's questions into a probe half and a scoring half shared across models, fit every rule only on the training models' probe half, and measure each rule's accuracy gap to the hindsight best of the eight architectures on the held-out model's scoring items (20 splits).  The strongest no-probe baseline, deploying each task's training-best architecture, has a 1.59-point gap.  Single-probe rules derived from the decomposition do not beat it: routing a Star probe by the net oracle gap, by coverage against a fitted $O^\star$, or by both gives 1.55, 4.02, and 2.46 points, the first a statistical tie and the other two significantly worse.  A leak-free pilot over all eight architectures at $N=10$, deploying the winner at $N=30$, reaches 1.22 points (sd 0.26), below the baseline across the 20 paired splits (paired $t=5.2$).  Restricting the pilot to one archetype per behavior (Proposer-Critic and Chain) reaches 0.94 (sd 0.16) but selects that menu with hindsight.  The choice is therefore resolvable by a labeled pilot but not predictable from a cheap probe, consistent with a transform already near-optimal given its proposals.

\begin{table}[t]
\centering\small\setlength{\tabcolsep}{6.0pt}
\begin{tabular}{rrrr}
\toprule
Calibration $B$ & Comparable cells & Agree & Agreement \\
\midrule
3 & 196 & 149 & 76.0\% \\
5 & 199 & 177 & 88.9\% \\
7 & 199 & 185 & 93.0\% \\
10 & 198 & 190 & 96.0\% \\
\bottomrule
\end{tabular}
\caption{Held-out-budget stability of the oracle-crossing label.  The label $O<O^\star$ (final accuracy above proposal coverage) at calibration budget $B$ predicts the corresponding label at $N=30$. Cells tied at either endpoint are excluded.  Because the boundary exactly encodes the sign of $Y-O$ at each budget, this evaluates label persistence rather than an independently fitted classifier.}
\label{tab:boundary-stability}
\end{table}

For architectures with a positive extensive term, the aggregate dividend realization $\eta=\Delta Y/\mathcal E$ is 33.0\% (Star), 38.7\% (Persona-Star), 108.1\% (Proposer-Critic), 25.6\% (Tournament), 53.9\% (Tree), and 35.1\% (Diamond).  We leave $\eta$ undefined for the chains: their extensive terms are numerically near zero, so a ratio is unstable and their intensive behavior is already the informative description.

Per-cell decompositions with endpoint $O,s,g$ values, closure errors, the Persona-Star contrast, and pooled response parameters are released.

\section{Per-model level accuracy}

Tables~\ref{tab:accuracy} and~\ref{tab:code} report team accuracy at $N=30$ for every architecture and model, split by task category for QA and given separately for code.  They carry the per-model detail that the main paper's gain figure averages over: no architecture wins across all five models in any category, and the within-model architecture spread is large on arithmetic (mean 14.0 points) but small on multiple-choice (mean 4.3) and on code.

\begin{table}[t]
\centering\small\setlength{\tabcolsep}{3.0pt}
\begin{tabular}{lrrrrr}
\toprule
Architecture & Lla. & Min. & Nem. & Q2.5 & Q3 \\
\midrule
\multicolumn{6}{l}{\emph{Arithmetic} (GSM8K, GSMHard)}\\
Star & 45.7 & 66.0 & 58.4 & 36.4 & 50.4 \\
Persona-Star & 45.0 & 69.0 & 61.3 & 37.1 & 56.7 \\
Prop.-Critic & \textbf{52.4} & 64.0 & \textbf{64.2} & \textbf{51.4} & \textbf{65.5} \\
Tournament & 28.3 & 62.0 & 58.6 & 43.0 & 54.6 \\
Tree & 41.5 & 65.8 & 61.8 & 46.1 & 52.4 \\
Diamond & 26.8 & 66.4 & 60.9 & 44.5 & 57.1 \\
Chain & 37.1 & 69.8 & 63.8 & 43.0 & 53.8 \\
Casc.-Chain & 35.1 & \textbf{69.9} & 57.8 & 40.9 & 52.5 \\
Single & 8.3 & 34.1 & 19.0 & 13.8 & 21.9 \\
\midrule
\multicolumn{6}{l}{\emph{Multiple-choice} (ARC, GPQA, MMLU)}\\
Star & 52.2 & 64.9 & 62.1 & 58.4 & 61.4 \\
Persona-Star & 53.0 & 64.5 & 62.0 & 59.5 & \textbf{63.3} \\
Prop.-Critic & 51.7 & 60.3 & 64.1 & 58.6 & 59.3 \\
Tournament & 49.4 & 64.0 & 62.3 & 59.1 & 62.1 \\
Tree & 53.1 & 64.0 & 61.4 & 59.6 & 61.1 \\
Diamond & 52.3 & 64.8 & 63.3 & \textbf{60.0} & 61.8 \\
Chain & \textbf{54.1} & \textbf{68.2} & \textbf{64.5} & 58.9 & 62.4 \\
Casc.-Chain & 53.7 & 66.2 & 62.0 & 58.2 & 61.7 \\
Single & 43.0 & 50.3 & 48.7 & 47.9 & 50.3 \\
\bottomrule
\end{tabular}
\caption{Team accuracy (\%) at $N=30$, equal-cell mean over each category's tasks, by model.  On arithmetic the architecture spread is large (mean 14 points) and Proposer-Critic leads four of five models, whereas on multiple-choice every design is within a few points (mean spread 4).  \emph{Single} is the mean individual worker, one non-thinking call.  Bold marks the best architecture within a model and category.}
\label{tab:accuracy}
\end{table}

\begin{table}[t]
\centering\small\setlength{\tabcolsep}{3.0pt}
\begin{tabular}{lrrrrr}
\toprule
Architecture & Lla. & Min. & Nem. & Q2.5 & Q3 \\
\midrule
Star & 58.9 & 66.1 & 84.3 & 69.3 & 78.3 \\
Persona-Star & 49.8 & 76.8 & 84.3 & 72.8 & \textbf{79.7} \\
Prop.-Critic & 54.3 & 59.8 & \textbf{84.6} & 70.9 & 77.6 \\
Tournament & 54.3 & 75.6 & \textbf{84.6} & 72.8 & \textbf{79.7} \\
Tree & \textbf{59.1} & 68.3 & \textbf{84.6} & \textbf{73.4} & 79.5 \\
Chain & 47.0 & 72.2 & 81.5 & 72.8 & 77.0 \\
Casc.-Chain & 51.8 & 70.3 & 81.7 & 73.0 & 77.6 \\
Diamond & 38.0 & \textbf{78.5} & 75.6 & 58.7 & 74.2 \\
\midrule
Single & 45.2 & 62.0 & 66.9 & 57.6 & 63.5 \\
\bottomrule
\end{tabular}
\caption{HumanEval+ pass rate (\%) at $N=30$ by architecture and model, with the mean individual-worker one-call baseline (Single).  Bold is the best architecture within a model.}
\label{tab:code}
\end{table}

\paragraph{Accuracy--cost frontier.}  Requested budget matches calls, not tokens.  At $N=30$ the equal-cell mean QA cost per problem ranges from 15.1k tokens for Tournament to 31.3k for Cascading-Chain, whose accumulating context grows fastest, a 2.1$\times$ spread at fixed $N$.  Read against final accuracy (the $Y$ column of Table~\ref{tab:transfer}), the non-dominated accuracy--cost set is Tournament (55.4\%, 15.1k), Star (56.4\%, 15.3k), Tree (57.3\%, 16.0k), and Proposer-Critic (59.1\%, 17.1k); Proposer-Critic is both the most accurate design overall and the most accurate point on that frontier.  Persona-Star, Chain, Diamond, and Cascading-Chain are each dominated by a cheaper design at equal or higher accuracy.

\section{Within-Task Structure and Task-Type Specialization}

The main-paper accuracies marginalize over items within a task.  Three cuts test whether that hides conclusion-flipping structure.  It does not, and one surfaces a result the aggregates omit.

\emph{Task-type specialization emerges with scale.}  Table~\ref{tab:math-by-n} tracks the Proposer-Critic arithmetic advantage against team size (the accuracy trajectories are shown in the main paper): below zero at $N=3$, indistinguishable from zero through $N=7$, and significantly positive from $N=10$ to $+5.7$ points over the runner-up at $N=30$, where it beats all seven other architectures with intervals excluding zero.  Because task identity is free, this refines the task-conditioned policy into an interpretable prior: on arithmetic, scale a Proposer-Critic team, while on the multiple-choice tasks the scaling gains are small for every design.

\begin{table}[t]
\centering\small\setlength{\tabcolsep}{4.5pt}
\begin{tabular}{rrlrr}
\toprule
$N$ & PC & runner-up & $\Delta$ & 95\% CI \\
\midrule
2 & 44.6 & Tree & -0.2 & $[-0.7, +0.2]$ \\
3 & 44.9 & Chain & -2.3$^\dagger$ & $[-2.9, -1.7]$ \\
5 & 49.0 & Chain & -0.0 & $[-0.6, +0.6]$ \\
7 & 51.7 & Persona-Star & +0.5 & $[-0.1, +1.1]$ \\
10 & 55.1 & Persona-Star & +2.8$^\ast$ & $[+2.2, +3.5]$ \\
15 & 56.1 & Persona-Star & +2.7$^\ast$ & $[+2.0, +3.4]$ \\
20 & 58.1 & Persona-Star & +4.4$^\ast$ & $[+3.8, +5.1]$ \\
30 & 59.5 & Persona-Star & +5.7$^\ast$ & $[+5.1, +6.4]$ \\
\bottomrule
\end{tabular}
\caption{Task-type specialization with scale on the two arithmetic word-problem benchmarks (GSM8K, GSMHard), equal-cell over 5 models $\times$ 2 tasks.  PC is Proposer-Critic accuracy (\%); runner-up is the best non-PC architecture at that budget (chosen on the full sample); $\Delta$ is their difference (points) with an item-clustered 95\% bootstrap interval (2{,}000 replicates resampling questions within benchmark).  $^\ast$/$^\dagger$ mark intervals excluding zero above/below.  PC is significantly behind the field-best at $N=3$ and significantly ahead from $N=10$; at $N=30$ it beats every other architecture (all seven intervals exclude zero, $+5.7$ to $+10.2$).}
\label{tab:math-by-n}
\end{table}

\emph{The pattern is not a pooled-subject artifact.}  Table~\ref{tab:mmlu-subject} splits MMLU-hard into its five constituent subjects: accuracies span only 54--62\% and Chain leads in four of five, so the no-universal-winner picture holds within the task rather than arising from averaging heterogeneous subjects.

\begin{table*}[t]
\centering\small\setlength{\tabcolsep}{4.0pt}
\begin{tabular}{lrrrrrrrrl}
\toprule
Subject & Star & Pers. & PC & Tourn. & Tree & Diamond & Chain & C.-Ch. & win \\
\midrule
college mathematics & 54 & 52 & 55 & 53 & 53 & 51 & 58 & 56 & Chain \\
college physics & 62 & 65 & 67 & 65 & 63 & 64 & 70 & 68 & Chain \\
econometrics & 56 & 56 & 55 & 56 & 57 & 56 & 58 & 56 & Chain \\
formal logic & 58 & 59 & 57 & 57 & 56 & 60 & 59 & 58 & Diam. \\
professional accounting & 57 & 59 & 55 & 57 & 57 & 58 & 59 & 57 & Chain \\
\bottomrule
\end{tabular}
\caption{MMLU-hard accuracy (\%) at $N=30$ by subject and architecture (pooled over 5 models).  The five hard subjects behave alike (accuracy 54--62\%) and Chain leads in four of five, so the no-universal-winner picture is not an artifact of pooling subjects.}
\label{tab:mmlu-subject}
\end{table*}

\emph{Architecture spread does not grow with problem difficulty.}  Binning GSM8K by the number of calculator steps in its annotated solution, and GSMHard by the same step count recovered through a digit-stripped content join to GSM8K (94.4\% of items matched), the spread across architectures widens with difficulty on GSM8K (from about 6 to 20 points) but is flat on GSMHard (7--9 points at every level).  The GSM8K pattern is a ceiling effect, since its easy items saturate near 80\%, rather than evidence that architecture matters more on harder problems.

\emph{Why the returns concentrate on arithmetic.}  The generate--transform decomposition locates the cause in its two margins (Star proposal tier, equal-cell over models, $N=3\to30$).  ARC coverage is already saturated ($O$ rises only from 89.6 to 91.2\%), so there is no extensive room.  GPQA and MMLU do add coverage ($O$ up 13.7 and 10.0 points), but the transform barely converts it ($\Delta Y=0.0$ and $+1.5$).  Only on arithmetic does coverage both grow and convert ($O$ up 23.9 and 14.8 points, $\Delta Y=+5.4$ and $+4.3$ for Star, larger for Proposer-Critic).  The task-averaged QA number superimposes these patterns, which is why it understates arithmetic and overstates the multiple-choice tasks.

\section{Controlled Intervention and Downstream Controls}

\subsection{Persona-Star intervention}\label{sec:persona-intervention}

\begin{figure*}[t]
\centering
\includegraphics[width=0.6\textwidth]{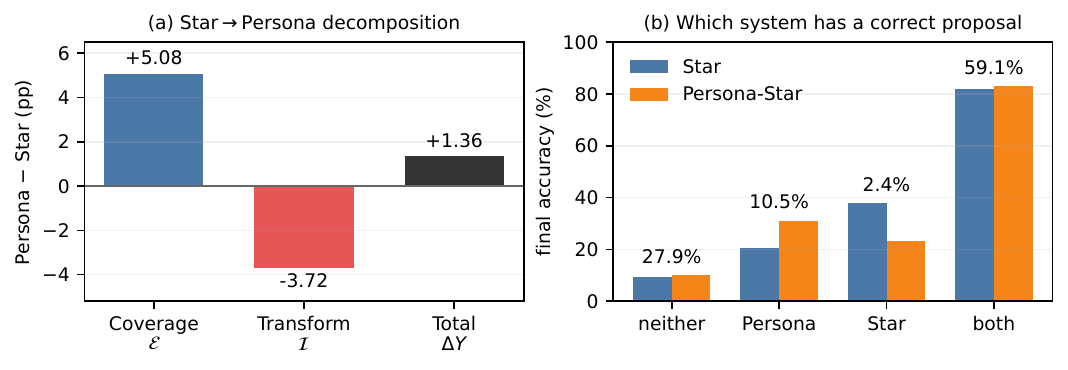}
\caption{Matched Star--Persona-Star prompt intervention at $N=30$.  (a) The intervention's two-margin decomposition: an extensive coverage dividend of $+5.08$ points is mostly offset by a $-3.72$ intensive transformation change, leaving $+1.36$ final points.  (b) Final accuracy split by which system has a correct proposal available (neither, Persona-Star only, Star only, or both), with equal-cell shares annotated.  The margin Persona newly covers converts weakly, and where both cover, Persona-Star is slightly more accurate, so the aggregate intensive term is not a same-item causal effect.}
\label{fig:persona-transfer}
\end{figure*}

The matched Star--Persona-Star comparison is a controlled prompt-only probe of the extensive margin, holding the graph and manager fixed and changing only the worker prompts.  Its two-margin decomposition (Figure~\ref{fig:persona-transfer}(a)) reads $\Delta Y=\mathcal E+\mathcal I$ with $\mathcal E=+5.08$ (the 8.13-point coverage gain weighted by midpoint leverage) and $\mathcal I=-3.72$ ($-3.50$ from covered-case success and $-0.22$ from recovery), giving $+1.36$ points.  The paired coverage transitions (Figure~\ref{fig:persona-transfer}(b)) localize the shortfall exactly as the $N=3\to30$ budget scaling does: the 10.53\% of trials Persona newly covers gain only $+0.94$ point, whereas on the 59.14\% both systems cover Persona is $+1.19$ points \emph{more} accurate (95\% CI $[0.41,1.98]$).  Newly created availability, not degradation on shared items, is what converts weakly.

\begin{table}[t]
\centering\small\setlength{\tabcolsep}{2.0pt}
\begin{tabular}{lrrrrr}
\toprule
Model & Star $A_{\rm pair}$ & Persona $A_{\rm pair}$ & $\Delta O$ & $\Delta\ell$ & $\Delta Y$ \\
\midrule
Llama-3.1 & 0.656 & 0.551 & +8.4 & +6.1 & +0.2 \\
Ministral-3 & 0.676 & 0.561 & +7.7 & +6.2 & +0.9 \\
Nemotron & 0.713 & 0.641 & +4.3 & +4.2 & +1.1 \\
Qwen2.5 & 0.823 & 0.685 & +12.9 & +9.2 & +0.9 \\
Qwen3 & 0.854 & 0.734 & +7.4 & +3.6 & +3.7 \\
\bottomrule
\end{tabular}
\caption{Matched Star--Persona-Star intervention at $N=30$, averaged over five QA tasks. The two agreement columns report exact-answer pairwise agreement.  The remaining columns are Persona-minus-Star percentage-point changes.}
\label{tab:persona}
\end{table}

\begin{table*}[t]
\centering\small
\begin{tabular}{llrr}
\toprule
Model & Task & $\Delta A_{\rm pair}$ & $\Delta$ accuracy (pp) \\
\midrule
Llama-3.1 & arc & -0.091 [-0.100,-0.082] & +0.8 [-0.3,+1.9] \\
Llama-3.1 & gpqa & -0.123 [-0.146,-0.102] & -0.5 [-3.7,+3.2] \\
Llama-3.1 & gsm8k & -0.098 [-0.107,-0.089] & -0.6 [-2.6,+1.2] \\
Llama-3.1 & gsmhard & -0.074 [-0.084,-0.062] & -0.8 [-2.4,+0.8] \\
Llama-3.1 & mmlu\_hard & -0.140 [-0.151,-0.128] & +2.1 [-0.0,+4.1] \\
Ministral-3 & arc & -0.085 [-0.092,-0.078] & -0.4 [-1.3,+0.5] \\
Ministral-3 & gpqa & -0.171 [-0.195,-0.148] & -1.3 [-5.6,+2.5] \\
Ministral-3 & gsm8k & -0.089 [-0.096,-0.083] & +3.1 [+1.6,+4.7] \\
Ministral-3 & gsmhard & -0.084 [-0.092,-0.076] & +2.9 [+1.0,+4.7] \\
Ministral-3 & mmlu\_hard & -0.147 [-0.158,-0.136] & +0.4 [-1.7,+2.4] \\
Nemotron & arc & -0.068 [-0.075,-0.060] & +0.1 [-0.6,+0.8] \\
Nemotron & gpqa & -0.087 [-0.107,-0.068] & -1.3 [-5.1,+2.0] \\
Nemotron & gsm8k & -0.046 [-0.052,-0.039] & +3.9 [+2.3,+5.5] \\
Nemotron & gsmhard & -0.067 [-0.076,-0.059] & +1.8 [+0.2,+3.4] \\
Nemotron & mmlu\_hard & -0.095 [-0.107,-0.083] & +1.0 [-0.6,+2.6] \\
Qwen2.5 & arc & -0.076 [-0.084,-0.067] & +0.8 [-0.3,+1.8] \\
Qwen2.5 & gpqa & -0.191 [-0.220,-0.164] & +2.9 [-0.8,+6.9] \\
Qwen2.5 & gsm8k & -0.127 [-0.138,-0.116] & +0.3 [-1.5,+2.2] \\
Qwen2.5 & gsmhard & -0.138 [-0.151,-0.125] & +1.0 [-0.7,+2.7] \\
Qwen2.5 & mmlu\_hard & -0.158 [-0.172,-0.145] & -0.2 [-2.1,+1.7] \\
Qwen3 & arc & -0.029 [-0.035,-0.023] & +0.5 [-0.4,+1.3] \\
Qwen3 & gpqa & -0.133 [-0.162,-0.104] & +3.7 [+0.5,+7.1] \\
Qwen3 & gsm8k & -0.151 [-0.163,-0.139] & +6.4 [+4.6,+8.3] \\
Qwen3 & gsmhard & -0.201 [-0.216,-0.185] & +6.1 [+4.0,+8.1] \\
Qwen3 & mmlu\_hard & -0.086 [-0.100,-0.073] & +1.6 [-0.4,+3.7] \\
\bottomrule
\end{tabular}
\caption{Paired Persona-Star minus Star effects at $N=30$. Brackets are item-clustered 95\% bootstrap intervals.}
\label{tab:persona-cells}
\end{table*}

\begin{table*}[t]
\centering\small\setlength{\tabcolsep}{3.2pt}
\begin{tabular}{lrrrrr}
\toprule
Quantity & Mean $\Delta$ (pp) & Aggregate 95\% CI & $\Delta>0$ & Cell CI$>0$ & Cell CI$<0$ \\
\midrule
Proposal oracle & +8.13 & [+7.50, +8.73] & 23/25 & 23 & 1 \\
Transfer loss & +5.85 & [+5.20, +6.49] & 20/25 & 19 & 2 \\
Recovery & -0.91 & [-1.13, -0.70] & 4/25 & 2 & 13 \\
Net transfer & -6.77 & [-7.45, -6.07] & 5/25 & 3 & 20 \\
Final accuracy & +1.36 & [+0.94, +1.82] & 18/25 & 7 & 0 \\
\bottomrule
\end{tabular}
\caption{Paired Persona-Star minus Star effects at $N=30$.  Sign counts summarize 25 item-clustered model--task intervals; the aggregate interval uses 5,000 benchmark-stratified item-bootstrap replicates, carries each sampled item across five models, and weights cells equally.  The means close $\Delta Y=\Delta O+\Delta r-\Delta\ell$.}
\label{tab:persona-transfer}
\end{table*}

\begin{table*}[t]
\centering\small\setlength{\tabcolsep}{4.3pt}
\begin{tabular}{lrrrrr}
\toprule
Coverage stratum & Share (\%) & Star $Y$ (\%) & Persona $Y$ (\%) & Cond. $\Delta Y$ (pp) & Contribution (pp) \\
\midrule
Neither covered ($0\!\to\!0$) & 27.94 & 9.49 & 10.12 & +0.63 & +0.08 \\
Persona only ($0\!\to\!1$) & 10.53 & 20.40 & 31.11 & +10.71 & +0.94 \\
Star only ($1\!\to\!0$) & 2.40 & 37.99 & 23.41 & -14.58 & -0.23 \\
Both covered ($1\!\to\!1$) & 59.14 & 81.80 & 82.99 & +1.19 & +0.58 \\
\midrule
Total & 100.00 & --- & --- & --- & +1.36 \\
\bottomrule
\end{tabular}
\caption{Paired coverage transitions for Persona-Star versus Star at $N=30$.  Each row is a realized $(O_{\rm Star},O_{\rm Persona})$ stratum.  Conditional $\Delta Y$ compares final accuracy on the same paired trials.  Contribution is stratum share times conditional $\Delta Y$, computed within each of the 25 model--task cells and then averaged, so it need not equal the product of the displayed aggregate columns.  It sums to the overall +1.36-point effect.}
\label{tab:persona-transitions}
\end{table*}

\begin{table*}[t]
\centering\small\setlength{\tabcolsep}{5.0pt}
\begin{tabular}{lrrrr}
\toprule
Architecture & $P(Y=0\mid O=1)$ & $A_{\rm pair}$: lost & preserved & $\Delta A_{\rm pair}$ \\
\midrule
Star & 19.6 & 0.590 & 0.793 & +0.202 (25/25) \\
Persona-Star & 25.5 & 0.504 & 0.690 & +0.186 (25/25) \\
\bottomrule
\end{tabular}
\caption{Proposal conflict and downstream loss at $N=30$, restricted to trials with an available correct proposal ($O=1$). Means weight the 25 model--task cells equally. The agreement gap is preserved minus lost and is positive in every cell.}
\label{tab:persona-conflict}
\end{table*}

Table~\ref{tab:persona} summarizes the matched intervention by model.  Table~\ref{tab:persona-cells} reports all 25 paired cells underlying the uniform agreement result.  Every agreement interval excludes zero in the negative direction, whereas only seven final-accuracy intervals exclude zero positively and none exclude zero negatively.  Table~\ref{tab:persona-transfer} reports both cellwise heterogeneity and the fixed-grid uncertainty of each equal-cell headline effect.  The 8.13-point proposal-coverage gain is positive in 23/25 cells, loss rises by 5.85 points, and recovery changes by $-0.91$.  Using unrounded means, $\Delta Y=\Delta O+\Delta r-\Delta\ell=1.36$ points.  Individual cell intervals are descriptive and are not treated as a multiplicity-corrected family of tests.

Table~\ref{tab:persona-transitions} separates composition from paired within-stratum performance.  Persona newly creates coverage on 10.53\% of trials and loses it on only 2.40\%, yielding the 8.13-point net coverage increase.  Yet the newly covered stratum contributes only +0.94 point to final accuracy.  On the 59.14\% of trials covered by both systems, Persona is +1.19 points more accurate, not worse.  The larger marginal Persona discard rate therefore partly reflects a harder covered population.

Table~\ref{tab:persona-conflict} conditions on correct-proposal availability.  Lost cases have lower agreement in all 25 Star and all 25 Persona cells.  The item-clustered interval for the agreement gap excludes zero in 24/25 Star and 25/25 Persona cells.  This within-system association is compatible with conflict making aggregation harder, but item difficulty remains an uncontrolled common cause.  Unlike the paired transition table, it does not identify why the two systems' marginal discard rates differ.

\subsection{Downstream transformation controls}

\begin{table*}[t]
\centering\small\setlength{\tabcolsep}{4.0pt}
\begin{tabular}{lrrrr}
\toprule
Final-manager instruction & $\Delta O$ & $\Delta\ell$ & $\Delta r$ & $\Delta Y$ \\
\midrule
Tally--decide & -0.02 [-0.15, +0.11] & +0.49 [+0.12, +0.87] & -2.67 [-2.85, -2.49] & -3.18 [-3.59, -2.78] \\
Critique--synthesize & +0.08 [-0.10, +0.26] & +0.27 [-0.12, +0.67] & -1.27 [-1.43, -1.11] & -1.46 [-1.87, -1.06] \\
\bottomrule
\end{tabular}
\caption{Diamond manager-prompt stress test at $N=30$.  Each entry is the prompt variant minus the baseline manager in percentage points, with a 95\% benchmark-stratified item-bootstrap interval.  The DAG, upstream instructions, model, sampling settings, and seed recipe are unchanged; separately launched decoding is not exact frozen-evidence replay.}
\label{tab:manager-prompt}
\end{table*}

\begin{table*}[t]
\centering\small
\begin{tabular}{lrrrr}
\toprule
Architecture & Coverage & Vote & Final agent & Final$-$vote \\
\midrule
Star & 61.5 & 45.9 & 56.4 & +10.5 \\
Persona-Star & 69.7 & 45.6 & 57.8 & +12.2 \\
Prop.-Critic & 59.1 & 45.8 & 59.1 & +13.3 \\
Tournament & 59.2 & 45.6 & 55.4 & +9.7 \\
Tree & 60.6 & 45.8 & 57.3 & +11.5 \\
Chain & 44.6 & 44.6 & 58.4 & +13.8 \\
Casc.-Chain & 44.5 & 44.5 & 56.7 & +12.2 \\
Diamond & 70.1 & 49.0 & 56.7 & +7.8 \\
\bottomrule
\end{tabular}
\caption{Active final-agent synthesis versus deterministic plurality over the same proposal tier at $N=30$ (equal mean over QA cells). Plurality is an offline control and invokes no additional LLM.}
\label{tab:vote}
\end{table*}

Table~\ref{tab:manager-prompt} tests two simple downstream interventions on Diamond.  Proposal coverage is stable: tally--decide changes $O$ by $-0.02$ points (95\% CI $[-0.15,0.11]$) and critique--synthesize by +0.08 ($[-0.10,0.26]$).  Nevertheless, final accuracy falls by 3.18 and 1.46 points, respectively.  Most of each decline is reduced recovery ($-2.67$ and $-1.27$ points), with only small loss increases (+0.49 and +0.27).  Thus neither generic instruction repairs the measured bottleneck.  Explicitly emphasizing selection or critique can suppress constructive synthesis.  Across paired runs, coverage status agrees on 97.7\% and 96.4\% of trials, while the complete proposal-answer vector agrees on 70.8\% and 60.0\%.  The stable aggregate coverage supports a downstream interpretation, but the lack of exact proposal replay prevents a stronger causal claim.

Table~\ref{tab:vote} recomputes exact plurality from the logged proposal answers.  Ties use the same MD5-derived item/run seed as the experiment.  For the serial designs plurality over $L=1$ is simply the initial worker.  Their final-minus-vote difference is therefore identical to net recovery from the initial proposal.  For all architectures, the control isolates what a no-LLM vote would obtain from the same first answer-producing tier.

\section{Direct Baseline and HumanEval+ Robustness}

\begin{table*}[t]
\centering\small\setlength{\tabcolsep}{3.0pt}
\begin{tabular}{lrrr}
\toprule
Architecture & Acc. & $\Delta$ vs.\ $N=1$ & $\Delta>0$ \\
\midrule
Star & 42.3 & +3.7 & 11/15 \\
Persona-Star & 43.6 & +5.0 & 11/15 \\
Prop.-Critic & 45.2 & +6.6 & 11/15 \\
Tournament & 41.6 & +3.0 & 9/15 \\
Tree & 42.8 & +4.3 & 10/15 \\
Chain & 44.8 & +6.2 & 12/15 \\
Casc.-Chain & 43.4 & +4.8 & 11/15 \\
Diamond & 43.3 & +4.8 & 11/15 \\
\midrule
Best per cell & 47.4 & +8.9 & 13/15 \\
\bottomrule
\end{tabular}
\caption{Fixed $N=30$ sparse architectures versus the direct one-call baseline on 15 hard-QA model--task cells.  Accuracy and change are equal-cell means in percentage points.  The final row is a descriptive post-hoc upper envelope, not a deployable policy.}
\label{tab:single-qa}
\end{table*}

Table~\ref{tab:single-qa} compares each fixed architecture to a purpose-built direct single-agent baseline, stronger than the ordinary single agent of Table~\ref{tab:accuracy} (it uses a cleaner prompt, and long reasoning for Nemotron).  Every architecture has a positive equal-cell mean, but only 9--12 of 15 individual cells improve, so the like-for-like lift over the ordinary single agent shrinks against this stronger one.  The best-observed row is selected after seeing the grid and is therefore an upper envelope, not a fair fixed-policy estimate.

\begin{table*}[t]
\centering\small\setlength{\tabcolsep}{4.5pt}
\begin{tabular}{llrrrrr}
\toprule
Model & Task & Items & Direct & Long-10$\times$ & $\Delta$ (pp) & Long tokens \\
\midrule
Llama-3.1 & gpqa & 198 & 26.8 & 24.7 & -2.0 [-8.1,+4.0] & 2.77k \\
Llama-3.1 & gsmhard & 1017 & 5.3 & 5.4 & +0.1 [-1.1,+1.3] & 5.15k \\
Llama-3.1 & mmlu\_hard & 724 & 46.1 & 45.4 & -0.7 [-3.5,+2.1] & 1.31k \\
Ministral-3 & gsmhard & 200 & 27.0 & 58.0 & +31.0 [+23.5,+38.0] & 1.97k \\
Ministral-3 & mmlu\_hard & 724 & 55.4 & 66.3 & +10.9 [+7.2,+15.1] & 1.92k \\
Nemotron & gpqa & 198 & 33.3 & 53.0 & +19.7 [+11.6,+27.8] & 5.29k \\
Nemotron & gsmhard & 1017 & 68.9 & 75.3 & +6.4 [+4.8,+8.2] & 2.05k \\
Nemotron & mmlu\_hard & 724 & 73.8 & 81.2 & +7.5 [+4.7,+10.2] & 1.72k \\
Qwen2.5 & gpqa & 198 & 30.3 & 30.3 & +0.0 [-4.0,+4.0] & 0.83k \\
Qwen2.5 & gsmhard & 1017 & 10.0 & 25.2 & +15.1 [+12.7,+17.4] & 0.84k \\
Qwen2.5 & mmlu\_hard & 724 & 55.5 & 53.7 & -1.8 [-3.6,-0.1] & 0.52k \\
Qwen3 & gpqa & 198 & 33.3 & 33.3 & +0.0 [-4.0,+4.0] & 1.85k \\
Qwen3 & gsmhard & 1017 & 15.5 & 50.3 & +34.8 [+31.9,+37.7] & 0.88k \\
Qwen3 & mmlu\_hard & 724 & 59.1 & 66.6 & +7.5 [+5.0,+10.2] & 0.83k \\
\bottomrule
\end{tabular}
\caption{One-agent 10$\times$-cap long-reasoning control on shared hard-QA items. The maximum generation length rises from 1,024 to 10,240 tokens and the prompt requests extended verification; hence this is a combined prompt-and-cap control, not a pure token intervention. $\Delta$ is Long minus Direct with a paired item-bootstrap 95\% CI. No Ministral--GPQA long-cap run is available.}
\label{tab:long-decode}
\end{table*}

Table~\ref{tab:long-decode} pairs the ordinary and long-reasoning single-agent runs on item id.  Equal weighting over its 14 model--task cells gives $+9.18$ points (fixed-grid, benchmark-stratified 95\% CI $[8.05,10.35]$), while observed prompt-plus-output tokens rise from 528 to 1,995 per item.  On exactly these items, every fixed $N=30$ sparse architecture has lower equal-cell mean accuracy than the long control.  The post-hoc per-cell sparse upper envelope is effectively tied ($-0.16$ points, 9/14 cell wins) while using 13.5$\times$ as many observed tokens on average.  This is evidence that the ordinary direct baseline understates a stronger single-agent alternative, not evidence that the generation cap alone causes the gain.

\begin{table*}[t]
\centering\small\setlength{\tabcolsep}{5.0pt}
\begin{tabular}{lrrrrrr}
\toprule
Architecture & $\mathcal{E}$ & $\mathcal{I}$ & $\Delta Y$ & $O$ & $g$ & $Y-O$ \\
\midrule
Star & $+6.34$ & $-4.38$ & $+1.95$ & 84.5 & 1.0 & $-13.09$ \\
Persona-Star & $+8.49$ & $-7.59$ & $+0.89$ & 90.3 & 0.0 & $-17.64$ \\
Prop.-Critic & $+7.70$ & $-7.46$ & $+0.24$ & 82.9 & 7.2 & $-13.50$ \\
Tournament & $+5.22$ & $-1.36$ & $+3.86$ & 83.0 & 9.0 & $-9.63$ \\
Tree & $+5.81$ & $-1.87$ & $+3.94$ & 83.8 & 4.4 & $-10.81$ \\
Chain & $-0.33$ & $+0.62$ & $+0.28$ & 71.9 & 20.5 & $-1.83$ \\
Casc.-Chain & $+0.85$ & $+0.41$ & $+1.26$ & 72.0 & 17.9 & $-1.06$ \\
Diamond & $+17.99$ & $-11.98$ & $+6.02$ & 84.3 & 0.6 & $-19.31$ \\
\bottomrule
\end{tabular}
\caption{Exact two-margin decomposition on the open-ended HumanEval+ task, equal-cell means over the five models.  $\mathcal{E}$, $\mathcal{I}$, and $\Delta Y$ are the requested $N=3\to30$ change in percentage points ($\Delta Y=\mathcal{E}+\mathcal{I}$ exactly).  $O$, generative recovery $g=P(Y{=}1\mid O{=}0)$, and $Y-O$ are at $N=30$.  Coverage rises as on QA, but $g$ nearly vanishes for the wide star-family and every architecture finishes below its proposal oracle ($Y<O$).}
\label{tab:code-margins}
\end{table*}

Table~\ref{tab:code-margins} applies the exact two-margin decomposition to HumanEval+, using per-candidate test execution for the proposal boundary.  The width structure matches QA: proposal-expanding designs post a positive extensive term offset by a negative intensive term, and the chains add essentially no coverage.  What differs is downstream.  Generative recovery $g$ is 0--1\% for the wide star-family and at most 9\% elsewhere, so no architecture finishes above its proposal oracle, unlike the QA chains.  On open-ended output the transform stage seldom synthesizes a correct program absent from its pool, leaving final accuracy close to coverage minus loss.

\section{Dense Debate and No-Peer Revision}

\begin{table*}[t]
\centering\small\setlength{\tabcolsep}{1.8pt}
\begin{tabular}{llrrrrrrr}
\toprule
Model & Task & Single & Self R3 & Debate R2 & Debate R3 & Debate$-$self (pp) & Cost / self & Cost / best sparse \\
\midrule
Llama-3.1 & gpqa & 26.8 & 31.8 & 28.3 & 29.0 & -2.8 [-8.1,+2.0] & 3.7$\times$ & 7.4$\times$ \\
Llama-3.1 & gsmhard & 5.3 & 31.1 & 22.9 & 27.6 & -3.4 [-5.5,-1.5] & 5.2$\times$ & 19.8$\times$ \\
Llama-3.1 & mmlu\_hard & 46.1 & 48.6 & 46.4 & 47.4 & -1.2 [-4.1,+1.5] & 5.1$\times$ & 12.3$\times$ \\
Ministral-3 & gpqa & 38.9 & 41.4 & 43.9 & 45.5 & +4.0 [-1.5,+9.1] & 2.5$\times$ & 8.3$\times$ \\
Ministral-3 & gsmhard & 26.2 & 59.9 & 62.4 & 59.3 & -0.6 [-2.8,+1.5] & 2.7$\times$ & 7.3$\times$ \\
Ministral-3 & mmlu\_hard & 55.4 & 67.1 & 66.0 & 64.5 & -2.6 [-5.0,+0.0] & 2.9$\times$ & 8.6$\times$ \\
Nemotron & gpqa & 33.3 & 32.3 & 35.4 & 33.8 & +1.5 [-2.0,+5.1] & 2.8$\times$ & 16.9$\times$ \\
Nemotron & gsmhard & 68.9 & 73.6 & 74.2 & 74.6 & +1.0 [+0.1,+1.9] & 3.8$\times$ & 24.3$\times$ \\
Nemotron & mmlu\_hard & 73.8 & 77.9 & 78.5 & 78.6 & +0.7 [-1.0,+2.3] & 3.6$\times$ & 17.9$\times$ \\
Qwen2.5 & gpqa & 30.3 & 31.3 & 31.3 & 31.8 & +0.5 [-3.5,+4.5] & 4.9$\times$ & 14.2$\times$ \\
Qwen2.5 & gsmhard & 10.0 & 28.2 & 20.6 & 26.4 & -1.9 [-3.8,+0.2] & 6.6$\times$ & 19.7$\times$ \\
Qwen2.5 & mmlu\_hard & 55.5 & 58.7 & 55.4 & 57.9 & -0.8 [-2.6,+1.2] & 6.0$\times$ & 18.8$\times$ \\
Qwen3 & gpqa & 33.3 & 33.3 & 32.3 & 32.3 & -1.0 [-2.5,+0.0] & 5.2$\times$ & 14.0$\times$ \\
Qwen3 & gsmhard & 15.5 & 37.0 & 26.7 & 33.6 & -3.3 [-5.5,-1.3] & 6.7$\times$ & 15.8$\times$ \\
Qwen3 & mmlu\_hard & 59.1 & 60.4 & 59.0 & 59.9 & -0.4 [-1.9,+1.1] & 6.3$\times$ & 13.2$\times$ \\
\bottomrule
\end{tabular}
\caption{Direct one-call accuracy and $N=30$ three-round controls.  In Self, each agent revises only its own prior answer; Debate exposes every agent to the other 29 reports.  Self R3 and both Debate columns use their common item support; the 95\% item bootstrap pairs the two round-3 outcomes.  Both cost ratios accumulate rounds 1--3. All 15 hard-QA cells are available.}
\label{tab:debate}
\end{table*}

Rows use requested $N=30$ and the three hard-QA tasks, with all 15 model--task cells complete.  Full-mesh Debate exposes each agent to the other 29 reports.  In Self, every agent instead revises only its own preceding answer.  Both run for three rounds and aggregate the 30 answers by plurality.

Debate round 3 beats the direct call by 8.3 points on average and in 14/15 cells, but the post-hoc sparse upper envelope is higher in 10/15.  Round 3 minus round 2 is positive/zero/negative in 11/1/3 cells.  Against the more diagnostic Self control, item-paired Debate round 3 is 0.69 point lower on average, is higher in only 5/15 cells, and consumes 2.5--6.7$\times$ as many cumulative tokens.  Peer exchange therefore does not explain the gross gain over one direct call uniformly.  The final cost column divides cumulative Debate tokens through rounds 1--3 by those of the highest-accuracy observed sparse architecture.  Because that sparse choice is made after seeing the grid, it remains a descriptive upper envelope rather than a deployable router.

\paragraph{Code, full budget sweep.} On HumanEval+, Debate and Self additionally run the full sweep $N\in\{2,\dots,30\}$ for all five models, three rounds each, and the picture reverses in two ways.  First, one round of peer exchange captures Debate's entire benefit: equal-cell accuracy at $N=30$ is 73.6\% initial, 76.1\% after one round, and 76.1\% after two, so the second round adds nothing, unlike the 11/15 hard-QA cells where it helped.  Second, peer content now helps: at $N=30$ Debate beats the no-peer Self control by 2.4 points, positive for all five models, the opposite of the hard-QA result.  Even so, one-round Debate (60 calls) only ties the best sparse architecture (30 calls), winning on Llama and Ministral and losing on Nemotron, Qwen2.5, and Qwen3 (mean $+1.0$ point).  Dense peer exchange therefore helps more on executable code than on hard QA, but remains a roughly $2\times$-cost baseline that does not beat the sparse frontier.

\onecolumn
\section{Prompt Templates}
\label{sec:prompts}

Every node receives a task-specific template with a strict output contract. Braced placeholders are filled at run time: \texttt{\{question\}} (and \texttt{\{choices\}} for multiple choice), the parent reports \texttt{\{proposal\}}/\texttt{\{workers\}}/\texttt{\{reports\}}, and the counts \texttt{\{b\}}/\texttt{\{m\}}/\texttt{\{M\}}. Three output contracts appear, shown by the three worker templates below: multiple-choice (\texttt{FINAL: <A, B, C, or D>}), numeric (\texttt{FINAL: <integer>}), and code (one \texttt{python} fence preceded by \texttt{CONF}). For a given role the multiple-choice and numeric templates differ only in the task noun and the format block, so we reproduce the numeric template per role; the code template uses the code contract. The exact set for all three modalities is in the released code.

\paragraph{Worker, multiple-choice contract.}
\begin{lstlisting}
Answer the multiple-choice question. Be concise.

Question: {question}
{choices}

Format EXACTLY (answer and confidence FIRST, then reasoning):
FINAL: <A, B, C, or D>
CONF: <0-100>
RATIONALE: <brief reasoning, max 4 lines>
\end{lstlisting}

\paragraph{Worker, numeric contract.}
\begin{lstlisting}
Solve this math problem step by step. Be concise.

Problem: {question}

Format EXACTLY (answer and confidence FIRST, then reasoning):
FINAL: <integer>
CONF: <0-100>
RATIONALE: <brief reasoning, max 4 lines>
\end{lstlisting}

\paragraph{Worker, code contract.}
\begin{lstlisting}
{question}

Format EXACTLY:
CONF: <0-100>
SOLUTION:
```python
<full function definition here>
```
\end{lstlisting}

\paragraph{Personas (Persona-Star).} Each persona replaces only the worker's leading directive; the format block is the worker's. The six numeric directives:

\begin{lstlisting}
forward: Solve this math problem by working FORWARD from the given values step by step. Be concise.

backward: Solve this math problem by working BACKWARDS: propose a plausible numerical answer, then check whether it satisfies every constraint in the problem. Adjust until consistent. Be concise.

decomposer: Solve this math problem by first DECOMPOSING it into a sequence of simpler subproblems. Solve each subproblem in order; the answer to the last yields the final answer. Be concise.

stepback: Solve this math problem by first STEPPING BACK: identify the general method, formula, or theorem this problem requires, then apply it to the specific numbers. Be concise.

conservative: Solve this math problem CONSERVATIVELY: commit only after at least two independent verification passes (e.g. dimensional check, sanity bound, alternative derivation) agree on the answer. Report low CONF when checks disagree. Be concise.

contrarian: Solve this math problem as a CONTRARIAN. Arrive at an obvious answer, then deliberately attempt to find an error in your reasoning. If you find a real flaw, revise; otherwise report the original answer and note that your attempted falsification failed. Be concise.
\end{lstlisting}

On code the six personas are recast as coding strategies (leading line each):

\begin{lstlisting}
forward: Approach: implement the function FORWARD — translate the spec directly into code, handling each requirement in the order it appears in the signature and docstring.

backward: Approach: work BACKWARD from the examples — figure out the expected output for the docstring's example inputs first, then write code that reproduces them and generalizes to the rest.

decomposer: Approach: DECOMPOSE the task into 2-3 smaller steps (helper computations or sub-cases), solve each, then compose them into the final function.

stepback: Approach: STEP BACK first — name the general algorithm or data structure this calls for (sorting, hashing, a counter, two pointers, dynamic programming, ...), then implement that approach.

conservative: Approach: code CONSERVATIVELY — handle edge cases explicitly (empty input, zero, negatives, boundaries), mentally run the docstring examples before committing, and report lower CONF if any case is uncertain.

contrarian: Approach: as a CONTRARIAN, write the obvious implementation, then deliberately hunt for the input that breaks it (off-by-one, empty case, aliasing, overflow). If you find one, fix it; otherwise keep it and note the attack that failed.
\end{lstlisting}

\paragraph{Critic (Proposer-Critic).}
\begin{lstlisting}
You are a CRITIC. A worker produced the solution below. Find the flaws — check arithmetic, challenge assumptions, look for missed cases. If after critique the original answer still holds, say so explicitly and report it; otherwise report the answer your critique supports.

Problem: {question}

Worker solution to critique:
{proposal}

Format EXACTLY (your own answer after the critique):
FINAL: <integer>
CONF: <0-100>
RATIONALE: <list flaws or confirm soundness, max 4 lines>
\end{lstlisting}

\paragraph{Synthesizer (Tree).}
\begin{lstlisting}
You are a SYNTHESIZER. You have received {b} independent worker solutions to the same problem. Extract the most consistent calculation chain. Do not simply vote — verify the arithmetic of the answer you report.

Problem: {question}

Worker solutions (independent, no communication between them):
{workers}

Format EXACTLY:
FINAL: <integer>
CONF: <0-100>
RATIONALE: <integrated reasoning, verify the arithmetic, max 6 lines>
\end{lstlisting}

\paragraph{Refiner (Chain, Cascading-Chain).}
\begin{lstlisting}
You are a REFINER. The {b} previous worker(s) below solved this problem in sequence (oldest first). Read them, then produce your own solution. You may agree, disagree, or extend — but do not just copy.

Problem: {question}

Previous worker solutions (chronological):
{workers}

Format EXACTLY:
FINAL: <integer>
CONF: <0-100>
RATIONALE: <your refined reasoning, verify arithmetic, max 4 lines>
\end{lstlisting}

\paragraph{Duelist (Tournament).}
\begin{lstlisting}
You are a JUDGE in a tournament bracket. You will see solutions from {b} competitors. Compare them critically — check arithmetic, identify reasoning gaps — then commit to a single answer. If both agree, verify. If they disagree, pick the stronger solution and explain why.

Problem: {question}

Competitor solutions:
{workers}

Format EXACTLY:
FINAL: <integer>
CONF: <0-100>
RATIONALE: <which solution prevails and why, max 4 lines>
\end{lstlisting}

\paragraph{Duelist, bye (Tournament, one parent).}
\begin{lstlisting}
One contestant advanced via bye in this tournament round; their solution is below. Your job is to verify their arithmetic and logic, NOT pick between alternatives. If sound, report the same answer with your own confidence. If you find a flaw, report the answer your check supports.

Problem: {question}

Contestant solution:
{workers}

Format EXACTLY:
FINAL: <integer>
CONF: <0-100>
RATIONALE: <verification result; sound or flawed, with the load-bearing check, max 4 lines>
\end{lstlisting}

\paragraph{Planner (Diamond).}
\begin{lstlisting}
You are PLANNER #{i_one_indexed} of {M}. Your job is to describe ONE method for solving this math problem — do NOT solve it. The other {M_minus_one} planners are independently producing their own methods; your method should be DIFFERENT from the most obvious approach a solver would default to. Keep it short and specific (3-5 sentences). Do NOT output a final number.

Problem: {question}

Format EXACTLY:
APPROACH: <name of method, e.g., 'unit conversion then ratio', 'set up equation in x', 'work backwards from total'>
STEPS: <3-5 brief, actionable steps a solver should take, on one or two lines each>
\end{lstlisting}

\paragraph{Plan-solver (Diamond).}
\begin{lstlisting}
A planner has proposed the following method for this problem. Execute the method to arrive at the answer. You may deviate if you identify a flaw — but state explicitly when you do.

Problem: {question}

Plan:
{workers}

Format EXACTLY:
FINAL: <integer>
CONF: <0-100>
RATIONALE: <executed plan, note any deviations, max 4 lines>
\end{lstlisting}

\paragraph{Manager, baseline (all topologies).}
\begin{lstlisting}
You are the FINAL ADJUDICATOR. You will see {m} downstream reports (workers, refiners, duelists, plan-solvers, synthesizers, or critics depending on the team structure). You are explicitly empowered to OVERRIDE the majority if their arithmetic is wrong. Be skeptical. Re-derive if needed.

Problem: {question}

Downstream reports:
{reports}

Format EXACTLY:
FINAL: <integer>
CONF: <0-100>
RATIONALE: <audit-style reasoning, identify any computational errors, max 6 lines>
\end{lstlisting}

\paragraph{Manager stress-test variants (Appendix~\ref{sec:persona-intervention} is the Persona probe; these back the Diamond manager-prompt test of Appendix~G.2).} Same input format as the baseline manager, changing only the synthesis instruction.

\paragraph{Manager, tally-decide.}
\begin{lstlisting}
You are the FINAL ADJUDICATOR. You will see {m} downstream reports. Your decision MUST follow this two-step procedure:

STEP 1 — TALLY: Group the reports by their FINAL answer. Write the exact tally in your rationale (e.g., "42:3, 48:5, 50:1"). Identify the modal answer.
STEP 2 — DECIDE: If the modal answer is supported by sound arithmetic, commit to it. ONLY OVERRIDE the modal answer if you can identify a specific arithmetic or reasoning error in the supporting reports.

Problem: {question}

Downstream reports:
{reports}

Format EXACTLY:
FINAL: <integer>
CONF: <0-100>
RATIONALE: <Step 1 tally; Step 2 decision (modal or override with named error); max 6 lines>
\end{lstlisting}

\paragraph{Manager, critique-synthesize.}
\begin{lstlisting}
You are the FINAL ADJUDICATOR. You will see {m} downstream reports. Your decision MUST follow this two-step procedure:

STEP 1 — CRITIQUE: For each distinct FINAL answer present, identify ONE arithmetic or reasoning weakness in the supporting reports (or "checks out").
STEP 2 — SYNTHESIZE: Among the surviving (uncritiqued) candidate answers, commit to one. Re-derive if necessary; do not split the difference.

Problem: {question}

Downstream reports:
{reports}

Format EXACTLY:
FINAL: <integer>
CONF: <0-100>
RATIONALE: <one critique line per distinct answer; surviving decision; max 8 lines>
\end{lstlisting}

\end{document}